\PassOptionsToPackage{table}{xcolor}
\documentclass[]{xiaomiev}

\usepackage[toc,page,header]{appendix}
\usepackage{solarized-light}
\usepackage{array}
\usepackage{siunitx,array}
\usepackage{makecell,array}
\usepackage{framed}

\usepackage{bbding}
\usepackage{amsmath} 
\usepackage[colorinlistoftodos]{todonotes}
\usepackage{longtable}
\usepackage{hhline}
\usepackage{fancyvrb}
\usepackage{fvextra}
\usepackage{CJKutf8}
\usepackage{multicol}
\usepackage{tablefootnote}
\usepackage{threeparttable}
\usepackage{tabularx}
\usepackage{mdframed}
\usepackage{amsfonts}
\usepackage{mathtools}
\usepackage[usestackEOL]{stackengine}
\usepackage{booktabs}
\usepackage{siunitx}
\usepackage{makecell}
\usepackage[table]{xcolor}
\newcommand{\commentout}[1]{}
\renewcommand{\paragraph}[1]{\noindent\textbf{#1.}}

\usepackage{enumitem}

\setlist[itemize]{leftmargin=15pt}
\makeatletter
\providecommand{\@bottomtitlebar}{}
\makeatother

\RequirePackage{xspace}
\makeatletter
\DeclareRobustCommand\onedot{\futurelet\@let@token\@onedot}
\def\@onedot{\ifx\@let@token.\else.\null\fi\xspace}

\makeatother
\newcommand{\Ours}{\texttt{Xiaomi-Robotics-1}}

\title{Xiaomi-Robotics-1: 
 Scaling Vision-Language-Action Models \\ with over 100K Hours of Real-World Trajectories
}
\author[\textcolor{xiaomievblue}{1}]{Xiaomi Robotics}

\abstract{
We present \Ours{}, a foundational vision-language-action (VLA) model capable of (1) following diverse language instructions to perform a wide range of mobile manipulation tasks in unseen environments out-of-the-box, and (2) efficiently adapting to novel downstream tasks with minimal fine-tuning data.
We propose a two-stage training recipe consisting of pre-training and post-training.
During pre-training, we imbue the model with broad and generalizable action-generation capabilities by training on over 100k hours of real-world manipulation trajectories, collected via UMI devices across a massive scale of environments and tasks.
Crucially, we develop a scalable auto-labeling pipeline that annotates trajectory clips with natural languages describing scene state transitions, providing rich and precise conditioning for action learning.
During post-training, we aim to align these capabilities with robot embodiments and imperative task instructions that humans naturally use to prompt robots, effectively mapping descriptive state transition understanding into actionable task prompts.
Extensive experiments demonstrate strong scaling behavior.
\Ours{} consistently improves with increased data scales and model sizes during pre-training.
This scaling behavior directly transfers to post-training, where a stronger pre-training model yields better out-of-the-box performance in real-robot evaluations within unseen environments.
Furthermore, \Ours{} serves as a strong robot foundation policy that can be efficiently fine-tuned on complex, dexterous tasks with high data efficiency.
Across multiple simulation benchmarks, \Ours{} outperforms state-of-the-art methods.
Notably, it establishes a new state-of-the-art with a 57.4\% success rate on RoboCasa365, surpassing the previous best of 46.6\%.
Furthermore, it achieves an average score of 20.07 on RoboDojo, significantly outperforming the prior state-of-the-art (13.07).
Code and model checkpoints will be released.
Project page: \url{https://robotics.xiaomi.com/xiaomi-robotics-1.html}
}

\begin{document}
\maketitle
\footnotetext[1]{See Contributions section for full author list. Please send correspondence to \href{mailto:mi-robotics@xiaomi.com}{\mbox{mi-robotics@xiaomi.com}}.}

\section{Introduction}
\label{sec:introduction}

The remarkable capabilities of modern large models are fundamentally driven by scale, where massive and diverse training corpora have underpinned unprecedented leaps in performance for both large language models~\cite{kaplan2020scaling,hoffmann2022training,brown2020language,liu2024deepseek} and vision-language models~\cite{chen2022pali,team2023gemini,team2024gemini,achiam2023gpt}.
Recent work on vision-language-action (VLA) models~\cite{black2024pi_0,intelligence2025pi_05,intelligence2025pi_06star,intelligence2026pi_07,gr00tn1_2025,qwenrobotmanip,wu2026pragmatic} and world-action models (WAM)~\cite{ye2026world,li2026causal,yuan2026fast} has produced increasingly promising results in robot manipulation, with early evidence that policies become more capable and generalizable as the training data grows in scale and diversity.
Following the same scaling trajectory of large models is therefore a natural and appealing direction for robotics.
However, robotics is hindered by a unique bottleneck of data.
The dominant data collection paradigm, real-robot teleoperation, is slow, costly, and hardware-bound, making it difficult to scale.
Furthermore, teleoperated data tends to be highly redundant, concentrated on a narrow slice of tasks and environments, limiting the diversity of the data.

\begin{figure}[t]
    \centering
    \includegraphics[width=0.9\linewidth]{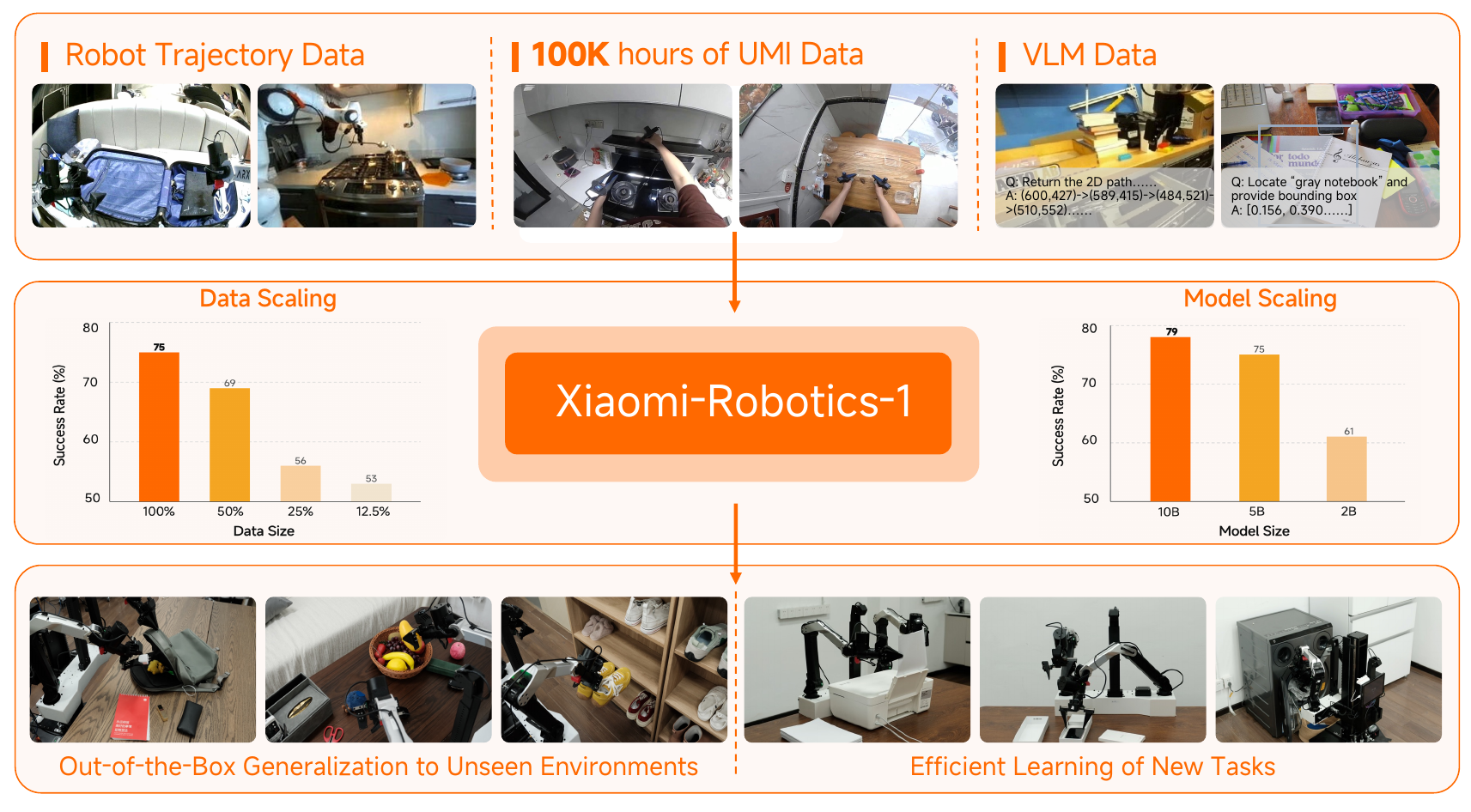}
    \caption{\textbf{Overview.} \Ours{} is pre-trained on over 100k hours of real-world UMI trajectories with auto-labeled state-transition language prompts.
    It is then aligned to robot embodiments and imperative instruction prompts via cross-embodiment post-training. \Ours{} scales effectively with data and model size. It is able to perform multiple tasks in unseen environment out-of-the-box and learn new tasks efficiently.}
    \label{fig:teaser}
\end{figure}

We present \Ours{} (Fig.~\ref{fig:teaser}), a foundational vision-language-action (VLA) model trained on a massive scale of real-world manipulation trajectories.
Drawing inspiration from the training paradigms of large language models, we propose a two-stage training recipe comprising pre-training and post-training.
During pre-training, we endow the model with robust and generalizable action-generation capabilities by leveraging data sources that scale readily in both volume and diversity.
Specifically, we curate a dataset of over 100k hours of real-world manipulation trajectories with UMI devices~\cite{chi2024universal}, spanning a wide range of environments and tasks.
Traditional trajectory labeling typically requires manual segmentation by task semantics and language annotations—a labor-intensive process that becomes prohibitive at this scale.
To address this challenge, we develop a scalable auto-labeling pipeline that leverages a pre-trained vision-language model (VLM)~\cite{qwen35blog} to annotate fixed-length trajectory segments with language descriptions detailing scene state transitions.
These annotations provide precise and sufficient semantic supervision.
Trained on these data, the model learns to generate actions that transform the scene from its current state to the language-specified target state (Fig.~\ref{fig:pretrain_visualization}).
In the post-training phase, we utilize over 10k hours of cross-embodiment data to align the strong action-generation capabilities acquired during pre-training.
This stage bridges two gaps: adapting the model from generating actions for UMI grippers to actions for robot embodiments, and transitioning from state-transition prompts to imperative instructions typically used by humans to prompt robots.
After post-training, \Ours{} is able to follow instructions and perform a wide range of tasks in unseen environments.
Furthermore, it serves as a strong robot foundation policy that can be efficiently fine-tuned to learn new tasks.

We perform extensive experiments to study the scaling properties of \Ours{}.
Results show that \Ours{} scales effectively during the pre-training phase, achieving lower validation action errors as data and model scale up.
Moreover, the scaling behavior observed in pre-training directly transfers to post-training, where stronger pre-training models yield better post-training success rates in out-of-the-box real-robot evaluation in unseen environments.
These results are encouraging, as they indicate that we are able to continue improving performance as we further scale data and model size.
In addition, we fine-tune the model on multiple complex dexterous tasks with minimal data.
\Ours{} achieves an average success rate of 75\% across four challenging tasks given less than 10 hours of data per task on average, outperforming $\pi_{0.5}$~\cite{intelligence2025pi_05} which obtains 40\%.
In addition, we evaluate \Ours{} on four challenging simulation benchmarks, \textit{i.e.}, RoboCasa~\cite{nasiriany2024robocasa}, RoboCasa365~\cite{nasiriany2026robocasa365}, VLABench~\cite{zhang2025vlabench}, and RoboDojo~\cite{chen2026robodojounifiedsimandrealbenchmark}.
\Ours{} achieves state-of-the-art results across all four benchmarks.
Notably, it sets a new state-of-the-art with a 57.6\% success rate on RoboCasa365, a substantial leap from the previous best of 46.6\%.
On RoboDojo, it delivers an average score of 20.07, significantly outperforming the prior state-of-the-art of 13.07.
Finally, \Ours{} enables the robot to autonomously accomplish a long-horizon, room-level mobile manipulation task of suitcase packing that spans over 10 minutes (see the project page for the video).
Code and model checkpoints will be released.
Project page: \url{https://robotics.xiaomi.com/xiaomi-robotics-1.html}
\section{Xiaomi-Robotics-1}
\label{sec:methods}
\Ours{} is an end-to-end vision-language-action (VLA) model trained at scale on heterogeneous data sources, including UMI trajectories, cross-embodiment robot trajectories, and vision-language data.
Given an observation $\mathbf{o}_{t}$ and a language instruction $l$, the model $\pi_{\theta}$ is trained to predict an action chunk $\mathbf{a}_{t:t+H}$ by maximizing the log-likelihood over the training dataset $\mathcal{D}$:
\begin{equation*}
    \max_{\theta} \mathbb{E}_{(\mathbf{o}_{t}, l, \mathbf{a}_{t:t+H})\sim \mathcal{D}} \log \pi_{\theta}(\mathbf{a}_{t:t+H}\mid\mathbf{o}_{t}, l)
\end{equation*}
We adopt a two-stage training recipe consisting of pre-training and post-training.
Pre-training leverages a scalable non-robot dataset with rich open-world diversity to endow the model with broad and generalizable representations for action generation.
Post-training then aligns these representations to robot embodiments and instruction-conditioned action generation, using a high-quality set of cross-embodiment data.
In the following sections, we describe the details of model architecture, data curation, and training recipe.

\begin{figure}[t]
    \centering
    \includegraphics[width=\linewidth]{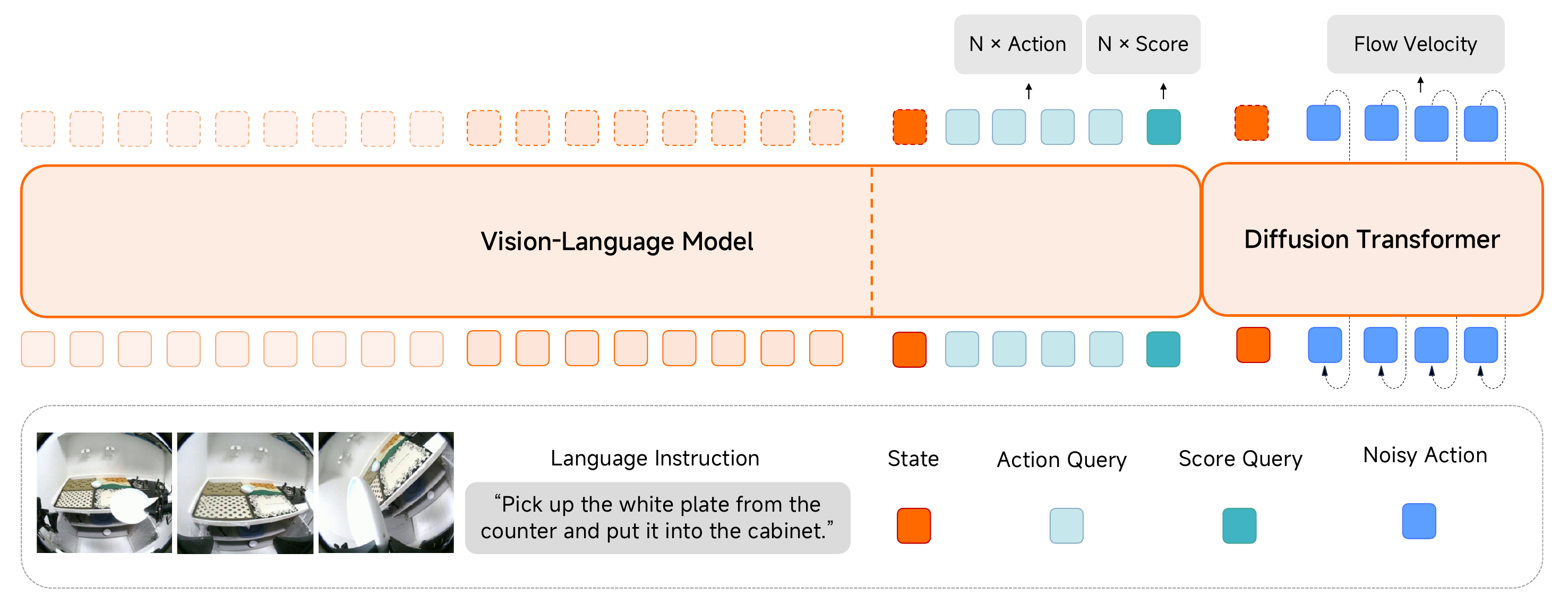}
    \caption{\textbf{Model Architecture.} \Ours{} adopts a Mixture-of-Transformers~\cite{liang2024mixture} architecture that couples a pre-trained VLM with a DiT. The VLM encodes the observation and language instruction, and additionally predicts action chunks via Choice Policies~\cite{qi2025coordinated} to accelerate training convergence. Conditioned on the robot state and the VLM's KV cache of the observation and language tokens, the DiT generates the action chunk via flow matching. Note that the action-related tokens from the VLM are excluded from the DiT's attention computation.}
    \label{fig:arch}
\end{figure}

\subsection{Model}
\label{subsection:model}
As illustrated in Fig.~\ref{fig:arch}, \Ours{} adopts a Mixture-of-Transformers (MoT)~\cite{liang2024mixture} architecture consisting of a pre-trained vision-language model (VLM) (\textit{i.e.}, Qwen3-VL~\cite{bai2025qwen3}) and a diffusion transformer (DiT)~\cite{peebles2023scalable}.
The DiT matches the VLM in the number of layers but employs a smaller hidden size for faster inference speed.
The model parameters for different scaling variants of \Ours{} are detailed in Tab.~\ref{tab:model_size}.
The VLM takes the current observation $\mathbf{o}_{t}$ and language instruction $l$ as inputs.
Conditioned on the robot proprioceptive state $\mathbf{s}_{t}$ and the KV cache produced by the VLM, the DiT generates the action chunk via flow-matching~\cite{liu2022flow}:
\begin{equation*}
    L_{\textrm{Flow}}(\theta) = ||\mathbf{v}_{\theta}(\mathbf{o}_{t}, l, \mathbf{s}_{t}, \tilde{\mathbf{a}}_{t:t+H}^{\tau}, \tau) - \mathbf{u}(\tilde{\mathbf{a}}_{t:t+H}^{\tau}, \mathbf{a}_{t:t+H}, \tau)||^{2}_{2}
\end{equation*}
$\tau$ is the flow-matching timestep.
$\tilde{\mathbf{a}}_{t:t+H}^{\tau} = \tau \mathbf{a}_{t:t+H} + (1 - \tau) \boldsymbol{\epsilon}$ is the noisy action where $\boldsymbol{\epsilon} \sim \mathcal{N}(\mathbf{0}, \mathbf{I})$.
Following~\cite{black2024pi_0}, we sample timestep $\tau$ from a Beta distribution, placing more weight on noisier timesteps during training:
\begin{equation*}
    u \sim \textrm{Beta}(1.5, 1), \quad \tau = (1-u) * 0.999 \in [0, 0.999]
\end{equation*}
Similar to~\cite{cai2026xiaomi}, we leverage adaptive normalization layers (adaLN)~\cite{peebles2023scalable} to inject the flow-matching timestep condition into the DiT for action generation.
During inference, we initialize the predicted action chunk from a random noise $\mathbf{a}^{\tau=0}_{t:t+H} \sim \mathcal{N}(\mathbf{0}, I)$.
The clean action chunk is recovered via a 5-step Euler integration, $\mathbf{a}^{\tau+\Delta \tau}_{t:t+H} = \mathbf{a}^{\tau}_{t:t+H} + \Delta\tau \cdot \mathbf{v}_{\theta}(\mathbf{o}_{t}, l, \mathbf{s}_{t}, \mathbf{a}_{t:t+H}^{\tau}, \tau)$, where the step size is set to $\Delta \tau=0.2$.

\begin{table}[t]
    \centering
    \caption{Model configurations for different scaling variants of \Ours{}.}
    \label{tab:model_size}
    \begin{tabular}{lccccccc}
        \toprule
        \multirow{2}{*}{Model} & \multirow{2}{*}{\# Layers} & \multicolumn{2}{c}{VLM} & \multicolumn{2}{c}{DiT} & \multirow{2}{*}{Total Params} \\
        \cmidrule(lr){3-4} \cmidrule(lr){5-6}
        & & Hidden Size & Params & Hidden Size & Params & \\
        \midrule
        \Ours{}-2B  & 28 & 2048 & 2.1B & 1024 & 470M & 2.6B \\
        \Ours{}-5B   & 36 & 2560 & 4.4B & 1024 & 604M & 5.1B \\
        \Ours{}-10B  & 36 & 4096 & 8.8B & 2048 & 1.5B & 10.5B \\
        \bottomrule
    \end{tabular}
\end{table}

To accelerate convergence~\cite{pertsch2025fast}, we introduce an auxiliary action-generation supervision on the VLM.
Specifically, we leverage Choice Policies~\cite{qi2025coordinated} to enable action generation directly within the VLM framework~\cite{cai2026xiaomi}.
We encode the robot state into a token using a multi-layer perceptron (MLP) and append it, along with the action and score query tokens, to the end of the vision-language token sequence.
The outputs corresponding to the action and score query tokens predict $K$ candidate action chunks and their associated $K$ scores, respectively.
We adopt a winner-takes-all paradigm as in~\cite{qi2025coordinated}, where only the candidate with the smallest $L_{1}$ loss is included in action loss computation:
\begin{equation*}
    L_{\textrm{Regression}}(\theta) = ||\hat{\mathbf{a}}^{*}_{t:t+H} - \mathbf{a}_{t:t+H}||_{1} + \sum_{k}^{K}||\hat{s}_{k} - s_{k}||_{2}^2
\end{equation*}
Let $\hat{\mathbf{a}}^{k}_{t:t+H}$ denotes the $k$-th predicted candidate action chunk, then $\hat{\mathbf{a}}^{*}_{t:t+H}$ is the one with the smallest $L_1$ distance to the ground truth $\mathbf{a}_{t:t+H}$.
$\hat{s}_{k}$ is the predicted score for the $k$-th candidate, and its regression target $s_{k}$ is defined as $s_{k} = ||\hat{\mathbf{a}}^{k}_{t:t+H} - \mathbf{a}_{t:t+H}||_{1}$.
That is, the $L_1$ distances between the $K$ predicted action chunks and the ground-truth action chunk serve as the target labels for score prediction.

Applying action-generation supervision directly on the VLM steers its representations toward features that better support action generation, thereby making the DiT learning more effective.
However, we empirically observe that letting the DiT tokens attend to the KV cache of these action-related tokens degrades performance.
We hypothesize that this arises from a shortcut in which the DiT simply copies the actions generated by the VLM rather than effectively grounding its own generation in the visual and textual context.
To mitigate this issue, we exclude these action-related tokens from the DiT's attention computation, constraining the DiT tokens to attend solely to the representations of the language instruction and visual observations.

\begin{figure}[t]
    \centering
    \includegraphics[width=\linewidth]{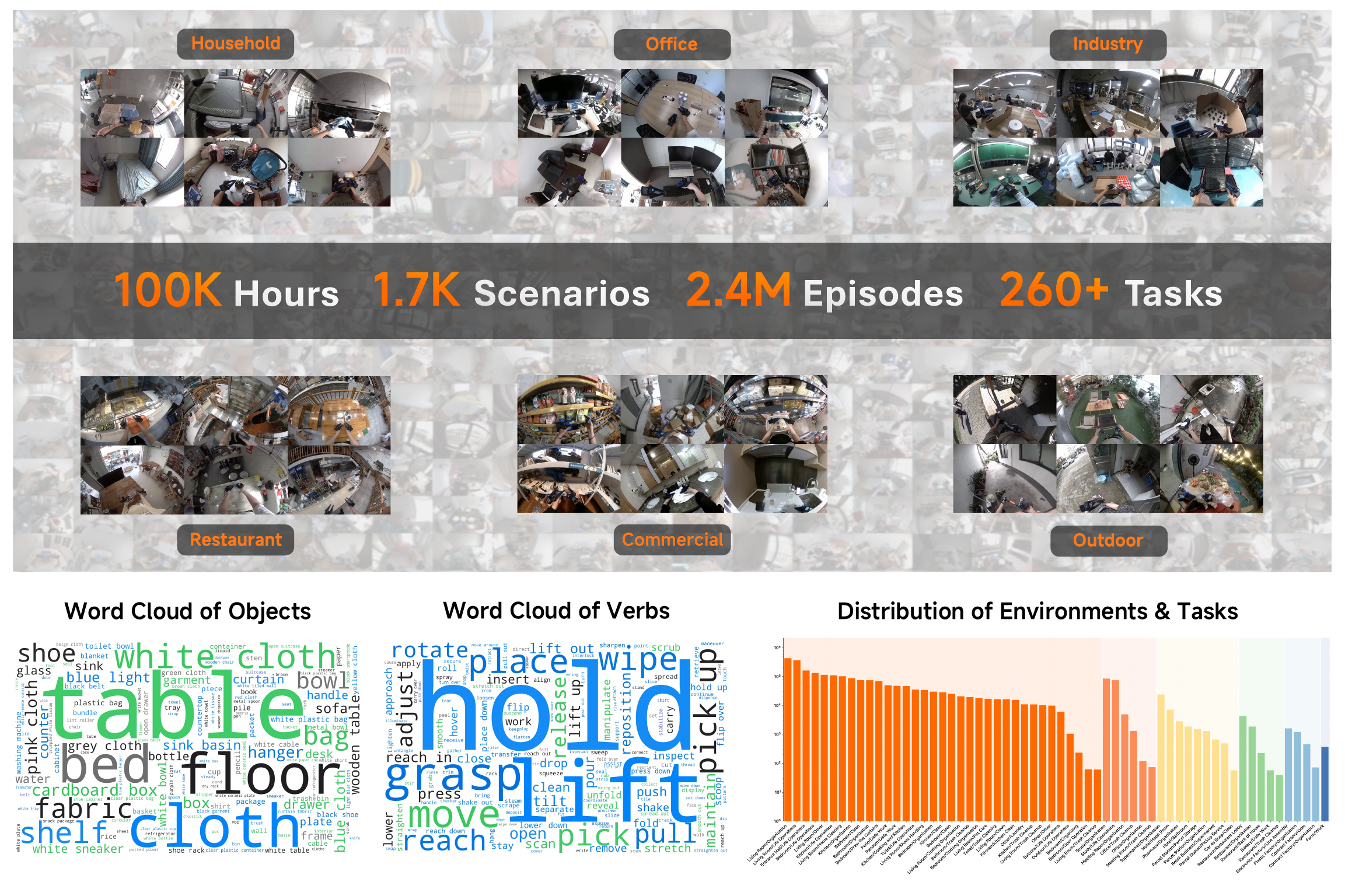}
    \caption{\textbf{Pre-training Dataset.} The pre-training dataset of \Ours{} contains over 100k hours of real-world manipulation trajectories collected with UMI devices.}
    \label{fig:pretrain_data}
\end{figure}

\subsection{Training \& Data}
\label{subsection:training}

\subsubsection{Pre-training}
During pre-training, our primary objective is to endow the model with broad and generalizable representations that transfer across diverse manipulation scenarios.
To this end, we curate a dataset of over 100{,}000 hours of real-world manipulation trajectories, captured with Universal Manipulation Interface (UMI) handheld grippers~\cite{chi2024universal} and egocentric cameras (Fig.~\ref{fig:pretrain_data}).
The dataset spans a diverse array of tasks collected across a massive scale of environments, including households, commercial premises, industrial sites, offices, and outdoor spaces.
Traditional robot trajectory annotation requires manually segmenting trajectories according to task semantics and labeling each segment with a language instruction---a labor-intensive process that becomes prohibitive at this scale.

To scale language annotation, we develop an auto-labeling pipeline that first divides each trajectory into equal-length segments and leverage Qwen3.5-27B~\cite{qwen35blog} to caption the state transitions of both the grippers and the interacting objects in the scene within each segment (see Fig.~\ref{fig:appendix_umi_pretrain} for examples).
To accelerate the annotation process, we develop a producer--consumer pipeline that decouples clip segmentations from caption labeling: while CPU worker threads cut per-segment clips into an in-memory filesystem, client threads keep hundreds of captioning requests in flight.
This highly effcient pipeline allows us to label the entire corpus of over 100k hours in roughly two weeks.
Trained on this dataset, the model learns to generate actions that drive the scene from the state in the current observation to the target state described by the language annotation.

The model is optimized to predict actions by jointly minimizing the flow-matching loss $L_{\textrm{Flow}}$ of the DiT and the regression loss $L_{\textrm{Regression}}$ of the VLM choice policy.
To preserve the vision-language capabilities of the pre-trained VLM, we further co-train the model on a high-quality vision-language dataset curated in our previous work~\cite{cai2026xiaomi} under the next-token prediction objective $L_{\textrm{NTP}}$.
The overall training objective is formulated as:
\begin{equation}
    L = L_{\textrm{Flow}} + L_{\textrm{Regression}} + \lambda L_{\textrm{NTP}}
\label{eq:loss}
\end{equation}
where $\lambda$ is set to $0.1$ in our experiments.
Vision-language data and UMI trajectories are sampled at a ratio of $1{:}9$.
To maximize training throughput, we pack all vision-language tokens within a batch into a single sequence for a VLM forward pass.
Since the VLM is computationally more expensive than the DiT, we amortize its cost by sampling four flow-matching timesteps per sample. 
The resulting four DiT inputs are similarly packed and processed in one DiT pass, conditioned on the corresponding unpacked VLM KV cache.

\begin{figure}[t]
    \centering
    \includegraphics[width=\linewidth]{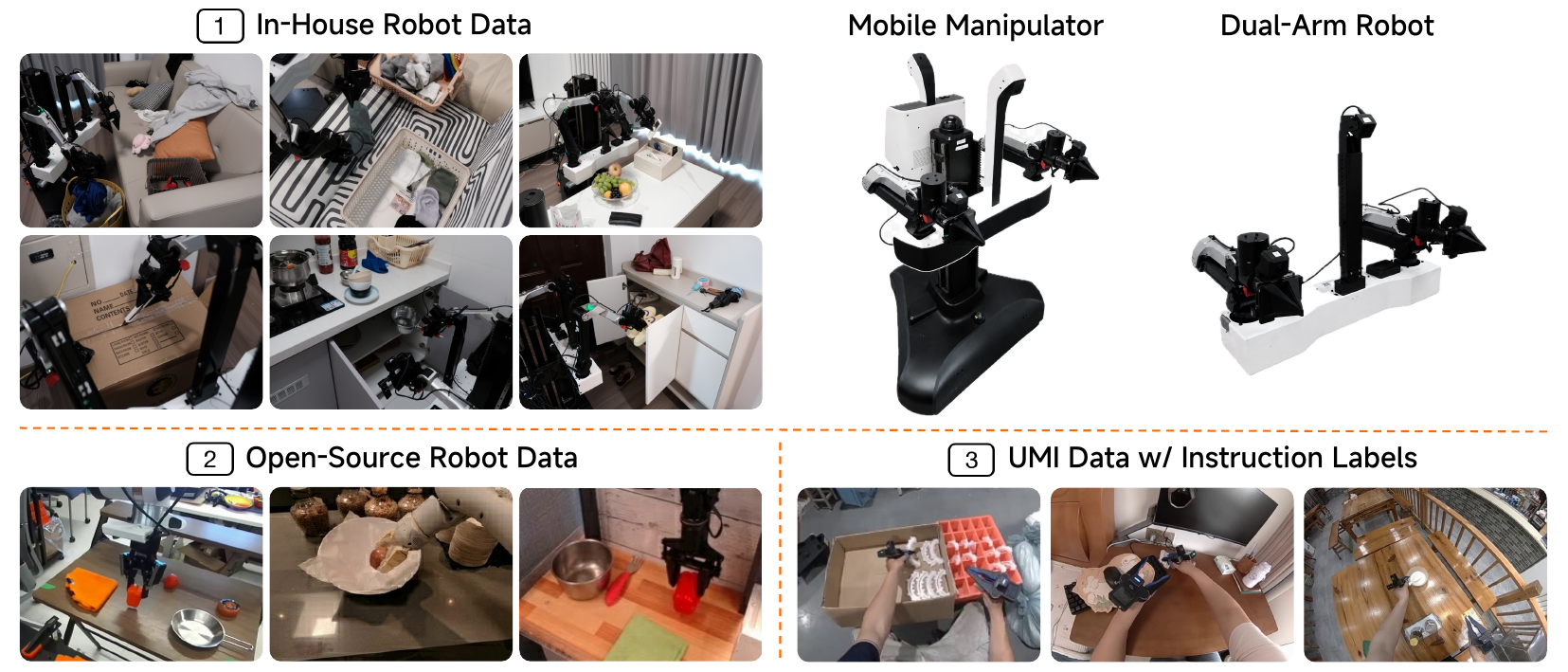}
    \caption{\textbf{Post-training Dataset.} The post-training dataset of \Ours{} comprises about 10k hours of cross-embodiment trajectories, including over 7.2k hours of in-house robot data collected with mobile manipulators and dual-arm robots, over 1k hours of instruction-labeled UMI data, and open-source robot datasets.}
    \label{fig:posttrain_data}
\end{figure}

\subsubsection{Post-training}
The goal of post-training is twofold.
First, we transfer the action-generation capabilities of UMI grippers acquired during pre-training to robot embodiments.
Second, we shift the language conditioning from the state-transition descriptions used in pre-training to the imperative instructions humans typically issue when prompting robots to perform tasks.

We curate the post-training dataset with cross-embodiment manipulation trajectories collected using UMI devices, static robot arms, and mobile manipulators.
Specifically, we collect over 7{,}200 hours of robot data using mobile manipulators and dual-arm robots across a diverse range of household environments and tasks (Fig.~\ref{fig:posttrain_data}).
We leverage Qwen3.5~\cite{qwen35blog} to annotate human-segmented video clips with language instructions.
In addition, we incorporate over 1{,}000 hours of human-annotated UMI data labeled with both temporal segments and language instructions.
Unlike the state-transition descriptions used in pre-training, these language instructions closely mirror how humans prompt robots to perform tasks, directly matching our alignment objective in the post-training phase (see Fig.~\ref{fig:appendix_umi_pretrain} and~\ref{fig:appendix_umi_posttrain} for comparison).
Finally, we include open-source robot datasets, including Bridge V2~\cite{walke2023bridgedata}, RT-1~\cite{brohan2022rt}, and DROID~\cite{khazatsky2024droid}.
We filter out idle segments within trajectories to prevent the model from learning uninformative or noisy signals.
In total, our post-training dataset comprises about 10{,}000 hours of trajectory data.

For arm actions, we adopt relative delta end-effector (EE) poses with respect to the current state:
\begin{equation*}
    \mathbf{a}_{t+i} = (\prescript{\textrm{EE}}{\textrm{Base}}{T}_{t})^{-1}\prescript{\textrm{EE}}{\textrm{Base}}{\hat{T}}_{t+i}
\end{equation*}
where $\prescript{\textrm{EE}}{\textrm{Base}}{T}_{t}$ denotes the pose of the end-effector with respect to the base at the current timestep $t$, and $\prescript{\textrm{EE}}{\textrm{Base}}{\hat{T}}_{t+i}$ represents the target end-effector pose at timestep $t+i$.
To align the arm action spaces across different embodiments, we unify the orientation of the end-effector frames across all robot data and UMI data in both the pre-training and post-training datasets.
Consequently, similar arm motions (\textit{e.g.}, moving forward or backward with respect to the end-effector frame) yield consistent action values regardless of the underlying hardware platform.
For mobile robot data, we represent the base and waist actions using the base velocity and the relative delta of the waist position, respectively.
To accommodate heterogeneous embodiments, we adopt a unified action vector for all trajectory data.
Although the arm actions are aligned across embodiments, the action spaces of different robots still differ in dimensionality.
We mask out the dimensions corresponding to missing action components during loss computation.

We train the model with the same objective as in pre-training (Eq.~\ref{eq:loss}).
Vision-language data, open-source robot data, instruction-labeled UMI data, and our in-house robot data are sampled at a ratio of $0.5{:}0.5{:}0.5{:}8.5$.
After post-training, the model can be prompted with language instructions to perform a wide range of tasks in unseen environments out-of-the-box.
In addition, it can efficiently adapt to novel downstream tasks with minimal amount of data.

\begin{figure}[t]
    \centering
    \includegraphics[width=0.95\linewidth]{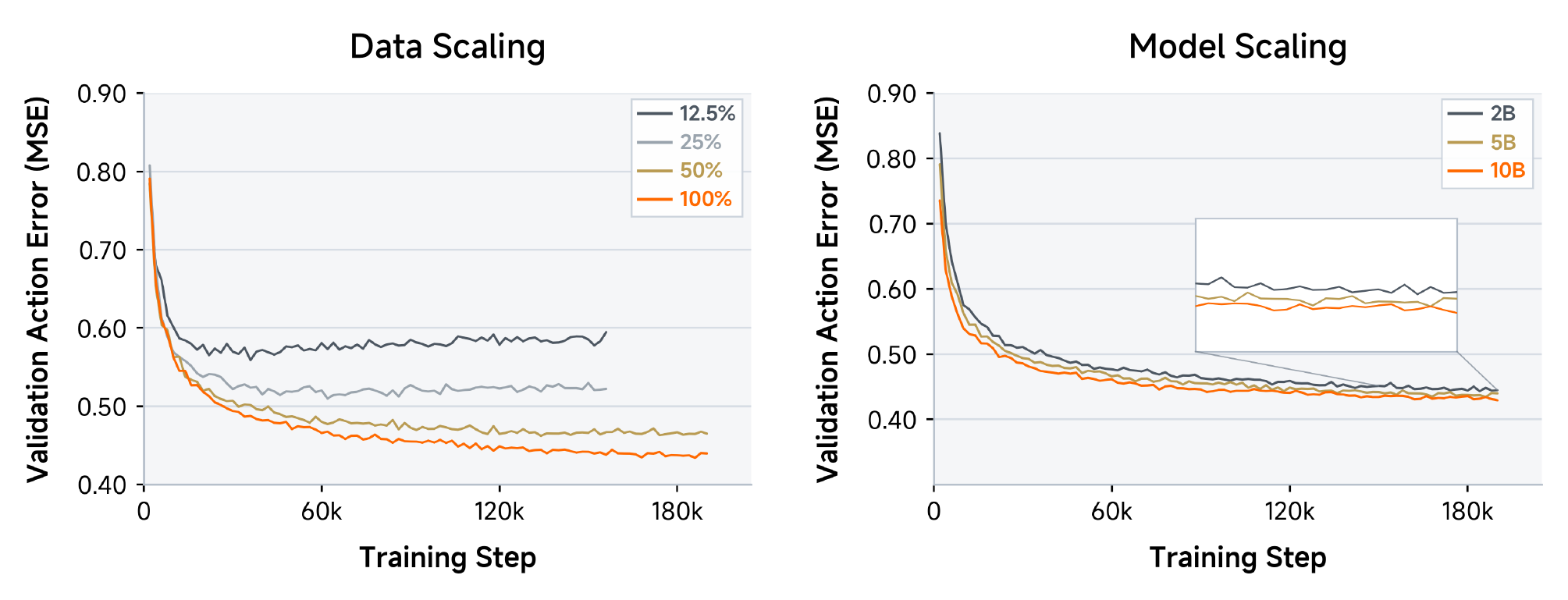}
    \caption{\textbf{Scaling of Pre-training.} We show the validation action errors (MSE) from the data-scaling and model-scaling pre-training experiments. We terminate the training for 12.5\% and 25\% data in the data-scaling experiment early as the validation loss indicates overfitting.}
    \label{fig:pretrain_scaling}
\end{figure}

\begin{figure}[t]
    \centering
    \includegraphics[width=0.95\linewidth]{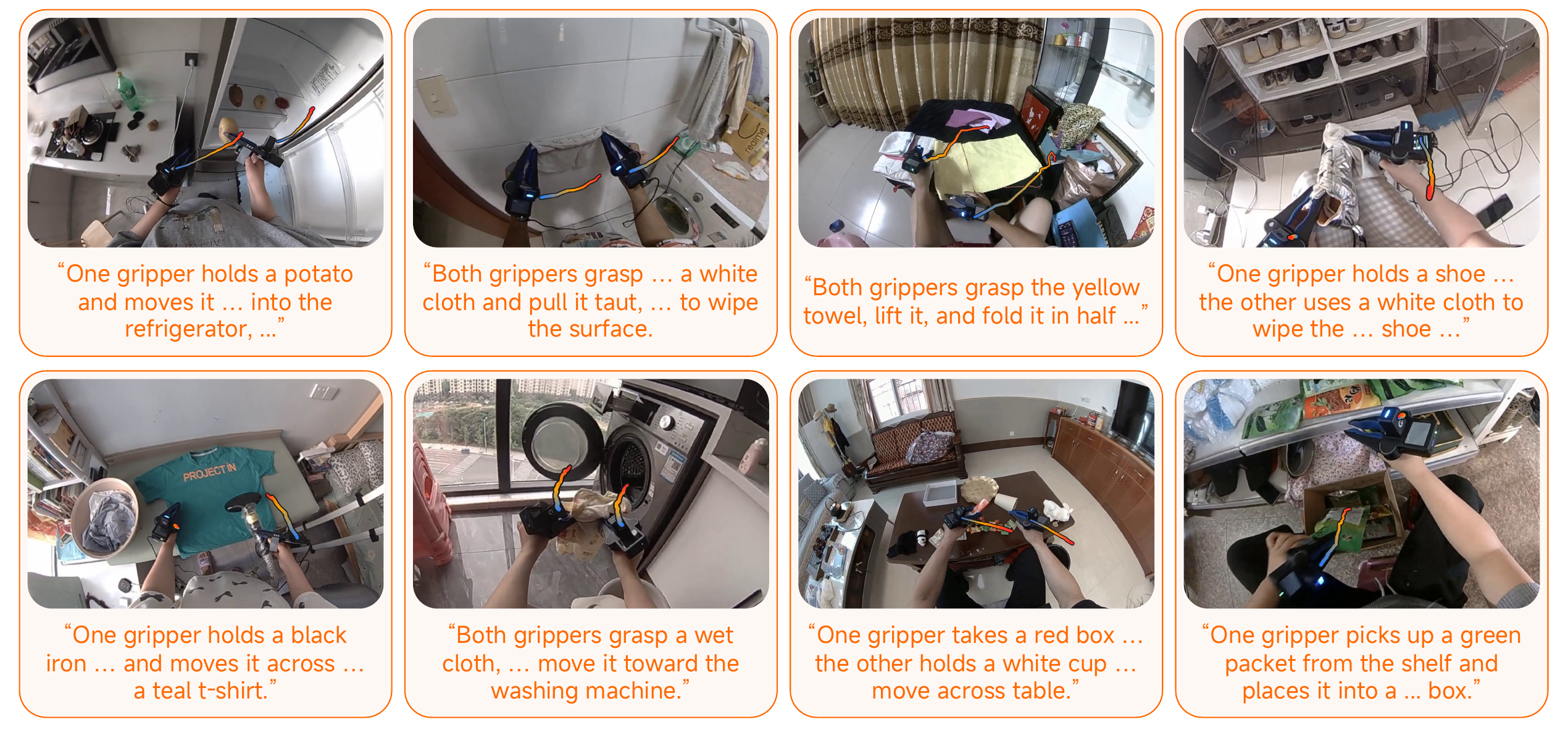}
    \caption{\textbf{Qualitative Results of Pre-training.} After pre-training, \Ours{} is able to predict action trajectories for UMI grippers on a held-out validation set according to the language description of state transitions.}
    \label{fig:pretrain_visualization}
\end{figure}

\section{Experiments}
\label{sec:experiments}

We design \Ours{} with \textit{scaling} in mind.
In this section, we investigate its scaling properties through extensive experiments.
Specifically, we design experiments to answer the following questions:
\begin{itemize}
    \item Does \Ours{} scale effectively with increasing data scale and model size during pre-training?
    \item Does a stronger pre-trained model translate to better post-training performance when evaluated out-of-the-box in novel environments?
    \item Can \Ours{} adapt to challenging new tasks with a minimal amount of data?
    \item How does \Ours{} compare to other robot foundation models in real-robot experiments and simulation benchmarks?
\end{itemize}

\subsection{Pre-training: Data and Model Scaling}
\label{subsec:experiments:pretrain}

\paragraph{Data Scaling}
We perform data-scaling experiments with \Ours{}-5B.
Due to compute budget limit, we pre-train the model on 12.5\%, 25\%, 50\%, and 100\% of about 20k hours of UMI data, respectively.
Each model is evaluated on a held-out validation set.
We use the mean-squared error (MSE) between the action predicted by flow-matching and the ground truth as the evaluation metric.
As shown in Fig.~\ref{fig:pretrain_scaling}, \Ours{} attains lower validation action errors with the increase of data scale.
With 12.5\% and 25\% of data, the validation action errors first decrease and then increase during training, indicating overfitting.
In contrast, the 50\% and 100\% data settings yield a monotonic decrease in loss, with the 20k setting exhibiting a steeper descent.
We show qualitative results of action prediction on validation data in Fig.~\ref{fig:pretrain_visualization}.

\paragraph{Model Scaling}
We perform model-scaling experiments on three size variants of \Ours{} (2B, 5B, and 10B) as specified in Tab.~\ref{tab:model_size}.
All three models are trained on the same 20k hours of data as in the data-scaling experiments and then evaluated on the same held-out validation set.
As illustrated in Fig.~\ref{fig:pretrain_scaling}, \Ours{} exhibits consistent improvements in action prediction precision as the model size scales up.
However, the performance gap among different model sizes are less pronounced than those observed across different data scales.
This result suggests that model capacity at the billions-parameter scale may already be sufficient to capture the current dataset's distribution, thereby making data volume the primary bottleneck for further generalization.
These findings do not diminish the value of model scaling, but rather highlight the critical importance of prioritizing the collection of large-scale, diverse datasets to unlock further performance gains.

\begin{figure}[!th]
    \centering
    \includegraphics[width=0.8\linewidth]{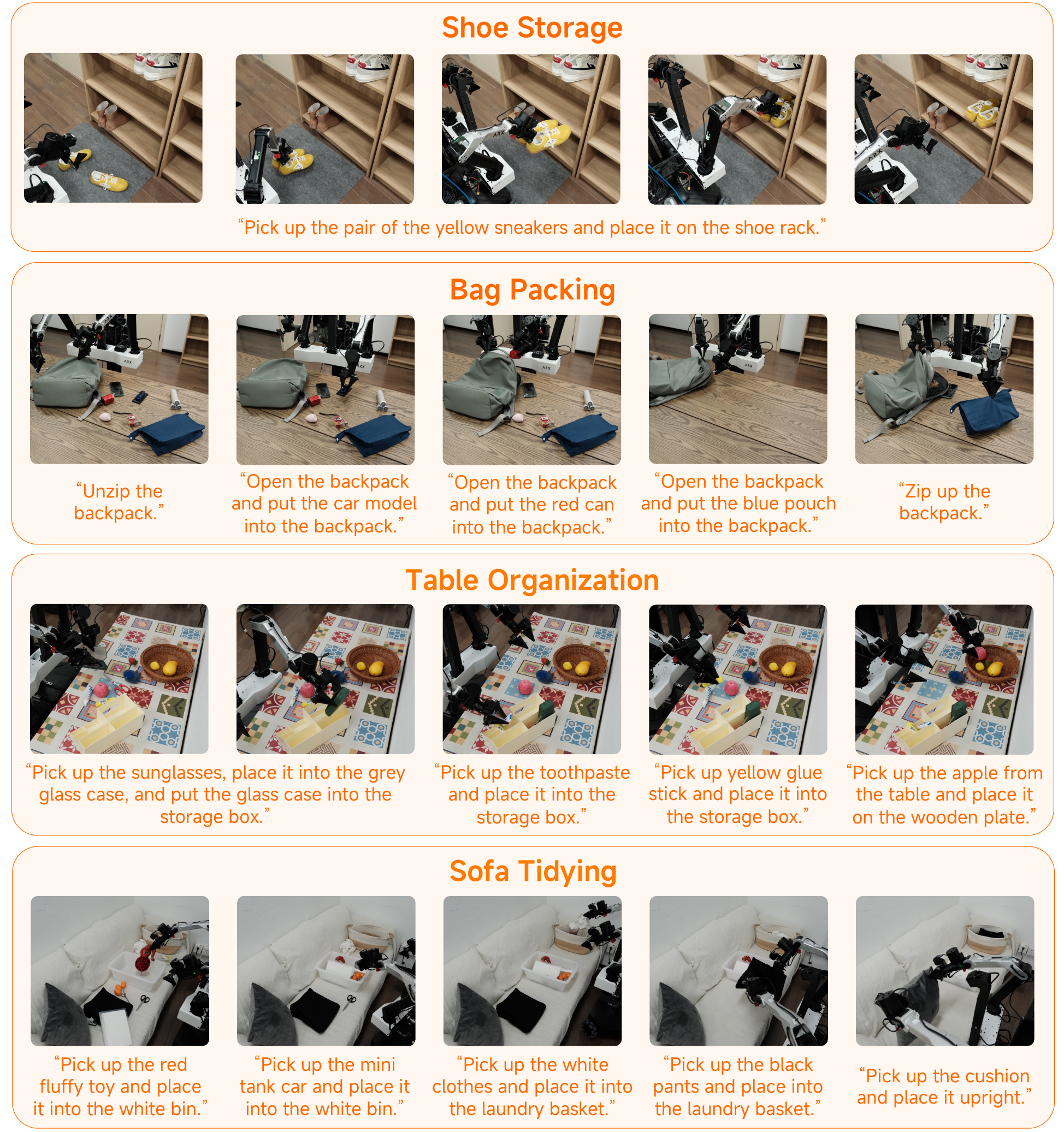}
    \caption{\textbf{Post-training Evaluation.} We evaluate the post-trained model out-of-the-box across four tasks in novel environments. Crucially, both the environments and object instances are unseen during training.}
    \label{fig:posttrain_experiment}
\end{figure}

\subsection{Post-training: Out-of-the-Box Evaluation in Novel Environments}
\label{subsec:experiments:posttrain}
In this section, we perform post-training experiments on the cross-embodiment post-training dataset and study its out-of-the-box performance in novel environments that are unseen during training.
In particular, we are interested in understanding whether the data scaling and model scaling properties from pre-training can transfer to post-training.
To mitigate overfitting, for the in-house robot data, we sample a diverse subset from the whole dataset for post-training.
Models are evaluated out-of-the-box in unseen environments after post-training without any per-task or per-environment fine-tuning.
Specifically, we evaluate on 4 tasks (Fig.~\ref{fig:posttrain_experiment}), \textit{i.e.}, shoe storage, bag packing, table organization, sofa tidying.
These tasks are seen in the post-training dataset but the environments and object instances during evaluation are unseen.

\subsubsection{Effectiveness of Scaling Pre-training Data}
\label{subsubsec:experiment:posttrain:data_size}
\begin{figure}[!th]
    \centering
    \includegraphics[width=0.95\linewidth]{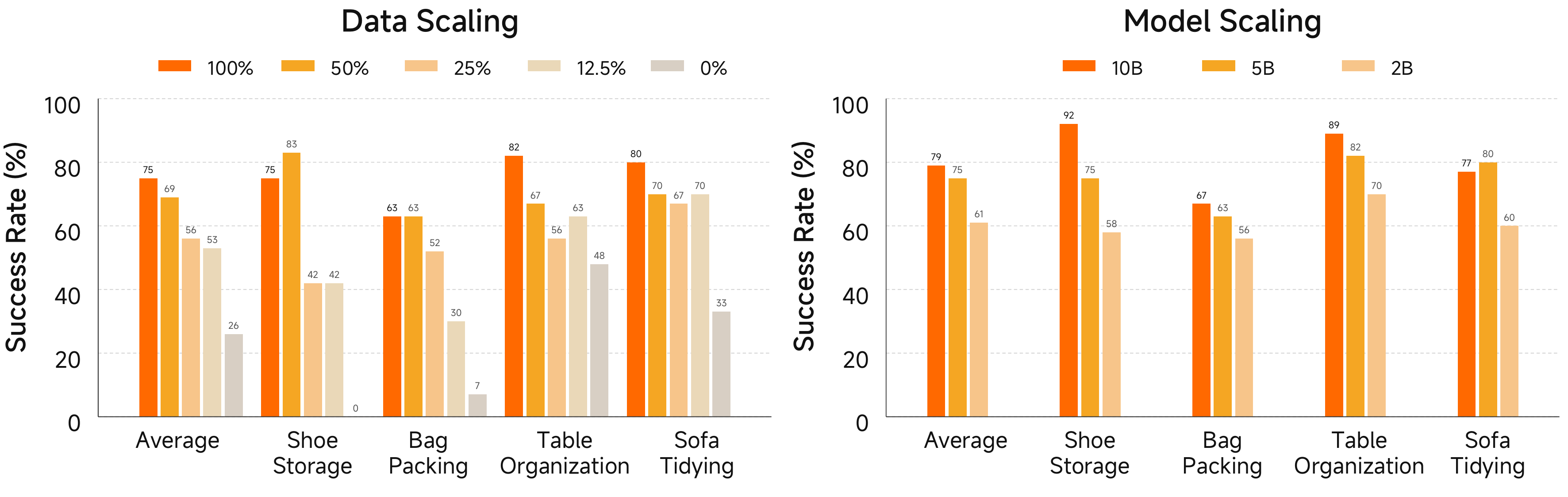}
    \caption{\textbf{Quantitative Results of Post-training.} We showcase the success rates of post-trained models across different pre-training data scales and model sizes.}
    \label{fig:posttrain_result}
\end{figure}

We first examine whether the benefits of scaling pre-training data transfer to post-training with the 5B variant of \Ours{}.
Using an identical training recipe, we post-train models initialized from checkpoints pre-trained on 12.5\%, 25\%, 50\%, and 100\% of 20k pre-training data (Sec.~\ref{subsec:experiments:pretrain}), alongside a baseline initialized from the Qwen3-VL~\cite{bai2025qwen3} pre-trained weight without any action pre-training.
Out-of-the-box evaluation results are shown in Fig.~\ref{fig:posttrain_result}.
The overall success rate increases monotonically with the scale of pre-training data, rising from 26\% without action pre-training to 75\% with 100\% of pre-training data.
The gains from scaling pre-training data are particular pronounced on tasks that demand contact-rich manipulation.
For instance, the baseline without pre-training fails completely on shoe tidying, whereas the model pre-trained on 100\% of the data reaches a 75\% success rate.
Notably, utilizing only 12.5\% of the pre-training data more than doubles the baseline's overall success rate (26\% vs. 53\%).
While the marginal gains gradually moderate as the pre-training corpus grows, the performance shows no sign of saturation: doubling the data from 50\% to 100\% yields an additional 6 percentage point improvement.
These findings suggest that further scaling robot pre-training data remains a highly promising avenue for achieving stronger out-of-the-box performance in unseen environments.

\subsubsection{Effectiveness of Scaling Model Size}
We further investigate the impact of model scale during post-training.
Specifically, We post-train the three size variants of \Ours{} (2B, 5B, and 10B) specified in Tab.~\ref{tab:model_size}.
These variants are initialized from checkpoints pre-trained on 20k hours of UMI pre-training data.
As shown in Fig.~\ref{fig:posttrain_result}, the overall success rate increases monotonically with model size, rising from 61\% for the 2B variant to 75\% and 79\% for the 5B and 10B variants, respectively.
Similar to data scaling, the gains from model scaling are most pronounced on shoe tidying, where the success rate climbs from 58\% (2B) to 75\% (5B) and further to 92\% (10B).
Performance improves consistently with model size across three out of four tasks, with the 5B and 10B variants performing comparably (80\% and 77\%) on sofa tidying.
Combined with the results in Sec.~\ref{subsubsec:experiment:posttrain:data_size}, these findings suggest that pre-training data scale and model size constitute two complementary axes for improving out-of-the-box performance in out-of-distribution settings.
And a stronger pre-trained model is able to translate to better out-of-the-box real-robot performance after post-training.

\subsection{Downstream Fine-tuning: Efficient Adaptation to New Tasks}
\label{subsec:experiments:finetune}

\begin{figure}[!th]
    \centering
    \includegraphics[width=0.8\linewidth]{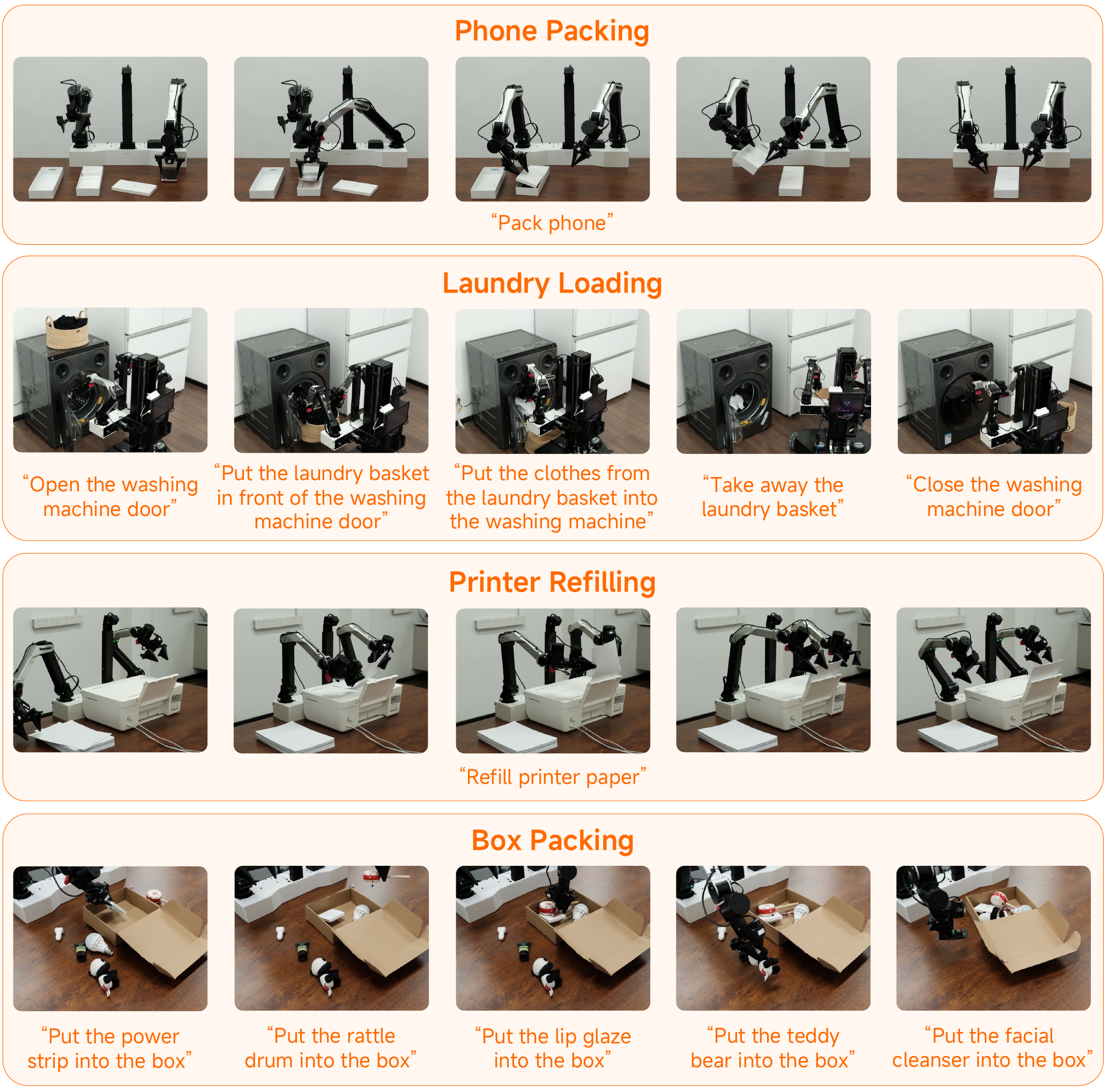}
    \caption{\textbf{Downstream Fine-tuning Evaluation.} We fine-tune the post-trained model on four new challenging tasks with a minimal amount of data.}
    \label{fig:finetune_experiment}
\end{figure}

A key desideratum of robot foundation models is their ability to efficiently adapt to novel tasks with minimal data.
To investigate how \Ours{} performs in this setting, we fine-tune our post-trained model on a suite of four novel challenging tasks: phone packing, laundry loading, printer refilling, and box packing (Fig.~\ref{fig:finetune_experiment}). 
Crucially, these tasks are entirely held out from the in-house robot dataset.
Each task introduces different complexities: 
phone packing requires bimanual coordination; 
laundry loading is a long-horizon mobile manipulation task involving multi-step instruction following; 
printer refilling demands handling of highly deformable sheets of paper; 
and box packing evaluates language grounding across multiple objects.

\begin{figure}[!th]
    \centering
    \includegraphics[width=0.95\linewidth]{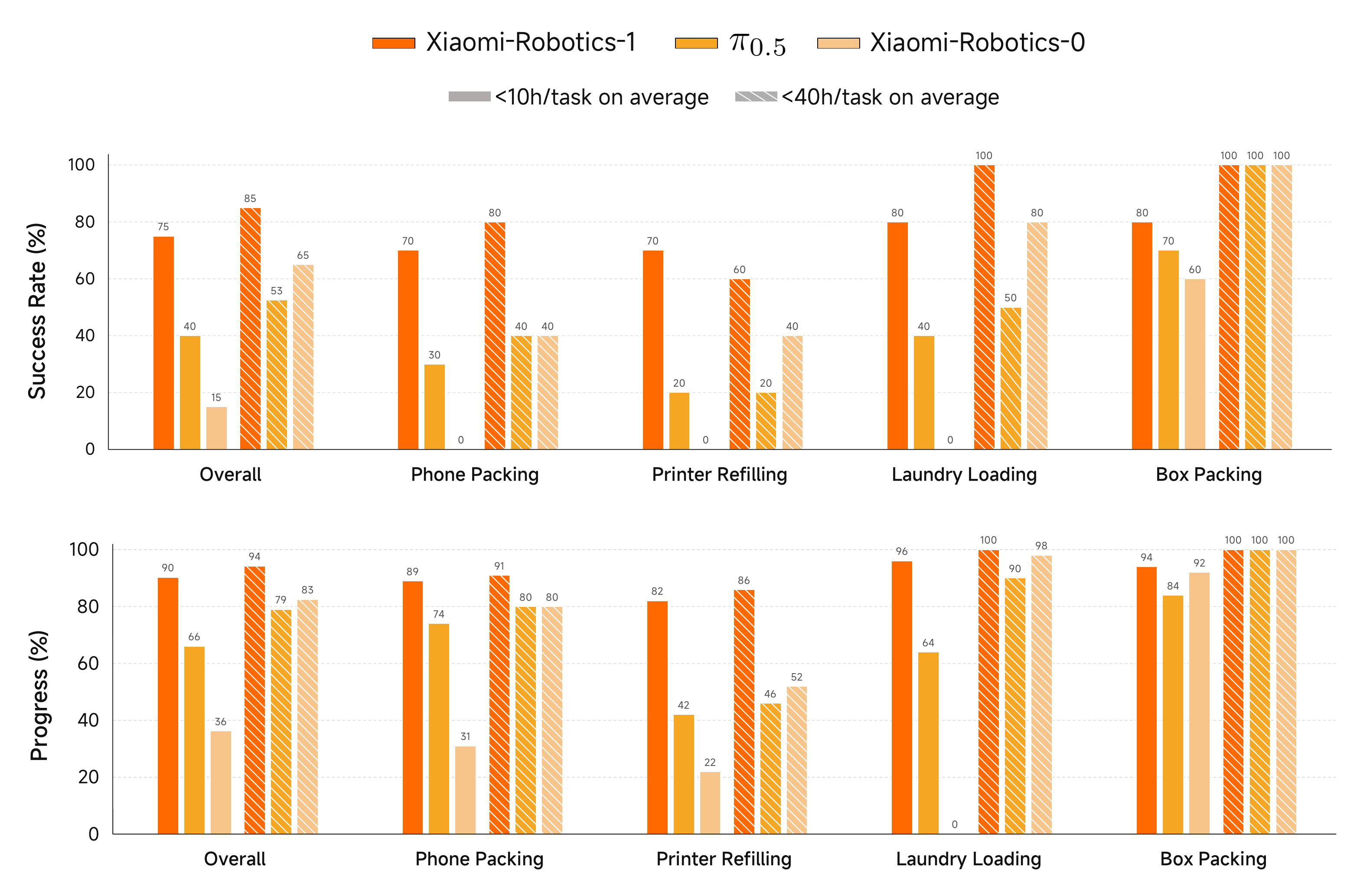}
    \caption{\textbf{Quantitative Results of Downstream Fine-tuning.} We report the success rates and progresses of different models across the four different tasks.}
    \label{fig:finetune_result}
\end{figure}

To evaluate data efficiency, we fine-tune the model under two settings.
For the high-data setting, we leverage a total of 144 hours of data across all tasks.
For the low-data setting, we sample 25\% of the data for each task from the high-data setting, resulting in a subset of 36 hours in total.
The average data per task for the low-data setting is less than 10 hours, with the maximum-data task, printer refilling, containing only 10.3 hours.
This poses a significant challenge for policy learning.
We fine-tune \Ours{} on the data of all four tasks using the asynchronous training recipe proposed in~\cite{cai2026xiaomi}.
We compare our method against two baselines: $\pi_{0.5}$~\cite{intelligence2025pi_05} and Xiaomi-Robotics-0~\cite{cai2026xiaomi}.
For $\pi_{0.5}$, we follow the official OpenPi\footnote{https://github.com/Physical-Intelligence/openpi} fine-tuning protocol and fine-tune the base model on these tasks.
For Xiaomi-Robotics-0~\cite{cai2026xiaomi}, we fine-tune its pre-trained model using the asynchronous setting.
Each model is evaluated for 10 trials per task.
We report both the average success rate and progress, where the progress measures partial task completion based on the task-specific milestones (Tab.~\ref{tab:finetune_progress}).
Results are shown in Fig.~\ref{fig:finetune_result}.

\Ours{} outperforms the two baseline methods in terms of average success rates and progress in both low-data and high-data settings.
In the low-data setting with less than 10 hours per task on average, it achieves an average success rate of 75\% and an average progress of 90\% across all four tasks, significantly outperforming $\pi_{0.5}$ with a 40\% success rate and a progress of 66\%.
The advantage of our method is most substantial on tasks requiring dexterous manipulation and mobile manipulation.
In phone packing, \Ours{} outperforms both baseline methods substantially in the two data settings.
In printer refilling, \Ours{} improves the success rate of the best baseline from 20\% to 70\% in the low-data setting, showcasing powerful capabilities in manipulating deformable objects.
In laundry loading, our method shows strong robustness over the full task horizon where failure may occur at any stage, achieving an 80\% success rate and a progress of 96\% in the low-data setting.
In this task, $\pi_{0.5}$ struggles with opening the washing machine while Xiaomi-Robotics-0 fails to complete the task in the low-data setting.
In box packing, all methods perform relatively well compared to other tasks, reaching 100\% success rates when more task-specific data are available.
Overall, all methods benefit from increasing the amount of fine-tuning data, but \Ours{} is substantially more data-efficient.
We attribute this advantage to large-scale pre-training which exposes the model to diverse environments and tasks, and careful post-training alignment that aligns the strong manipulation capabilities acquired during pre-training to robot embodiments and instruction prompts.
These results demonstrate that \Ours{} can serve as a strong foundation robot policy for efficient adaptation to novel tasks.

\subsection{Simulation Benchmarks}
\label{subsec:experiments:simulation}

In this section, we evaluate \Ours{} on four simulation benchmarks.
\begin{itemize}
    \item \textbf{RoboCasa}~\cite{nasiriany2024robocasa}:
    RoboCasa features single-arm manipulation in realistic kitchen environments.
    The benchmark contains 24 everyday kitchen manipulation tasks spanning pick-and-place, articulated-object interaction, appliance operation, and coffee-making.
    In order to test generalization capabilities, it evaluates policies on unseen object instances and includes two scenes of which the styles were unseen in the training data among the five evaluation scenes.
    For training, we use the official set of 300 synthetic demonstrations provided by the benchmark.
    Following the standard evaluation protocol, we report the average success rate over 100 evaluation episodes per task across the five evaluation scenes.
    Results are shown in Tab.~\ref{tab:robocasa}.
    
    \item \textbf{RoboCasa365}~\cite{nasiriany2026robocasa365}:
    RoboCasa365 extends RoboCasa into a large-scale simulation benchmark for evaluating general-purpose robot manipulation, with a particular emphasis on generalization beyond the training task distribution.
    It expands the original 24-task benchmark to 365 tasks across over 2,500 procedurally generated kitchen scenes and 3,200 object instances, covering both short-horizon manipulation and long-horizon mobile manipulation.
    This diversity introduces substantial variations in kitchen layouts, object appearances, and spatial relationships, testing whether policies remain robust across unseen scene and object configurations rather than overfitting.
    More importantly, RoboCasa365 explicitly evaluates generalization on task composition via a split featuring task templates that are unseen during training.
    We train our policy using the officially released dataset, which provides 100 demonstrations for each task.
    We follow the standard evaluation protocol and evaluate across 50 benchmark tasks, comprising 18 seen atomic tasks, 16 seen composite tasks, and 16 unseen composite tasks.
    The unseen composite tasks provide a zero-shot evaluation of whether the policy can recombine previously learned atomic skills and semantic knowledge to solve novel long-horizon tasks.
    Results are shown in Tab.~\ref{tab:robocasa365}.
    
    \item \textbf{VLABench}~\cite{zhang2025vlabench}:
    VLABench is a large-scale benchmark designed for comprehensive evaluation of language-conditioned manipulation.
    It comprises 100 task categories with over 2,000 object instances, emphasizing challenging scenarios involving semantic and distribution shifts. 
    Specifically, VLABench introduces instructions with implicit human intentions, long-horizon tasks requiring multi-step reasoning, and evaluation settings that demand commonsense reasoning, category-level generalization, and robustness to unseen object appearances.
    These challenges are distributed across five evaluation tracks—In-distribution, Cross-Category, Commonsense, Instruction, and Texture—to thoroughly assess policy generalization.
    Beyond success rate (SR), VLABench introduces progress score (PS) and intention score (IS) to measure task completion quality and instruction understanding.
    Following the standard evaluation protocol, we train our policy using only the official training set from the In-distribution track, which contains 10 tasks with 500 demonstrations per task.
    We leverage chain-of-thought (CoT) labeling as in ERVLA~\cite{sun2026revisiting} and train our model with a 50\% probability on the next-token-prediction loss of CoT alongside the action loss.
    During evaluation, each task is tested across all five tracks with 50 evaluation episodes per track, resulting in 250 rollouts for each task and 2,500 rollouts in total.
    Results are shown in Tab.~\ref{tab:vlabench}.
    
    \item \textbf{RoboDojo}~\cite{chen2026robodojounifiedsimandrealbenchmark}:
    RoboDojo is a unified simulation and real-world benchmark for comprehensively evaluating generalist robot manipulation policies. 
    Unlike existing benchmarks that focus primarily on individual skills, RoboDojo provides a diverse and challenging suite comprising over 42 simulation tasks with varied object configurations, scene layouts, and language objectives.
    These tasks are organized into five core capability dimensions: Generalization (robustness to object/layout changes), Precision (fine-grained control), Long-Horizon (multi-step execution), Memory (state-dependent manipulation), and Open (open-ended instruction following).
    Following the official protocol, we evaluate \Ours{} on the RoboDojo simulation benchmark and report the average score and success rate of each capability dimensions.
    Results are shown in Tab.~\ref{tab:robodojo}.
\end{itemize}

\begin{table}[!t]
\centering
\small
\caption{\textbf{Results on the RoboCasa Benchmark.} We report the average success rate (\%). The best and second-best results are highlighted in \textbf{bold} and \underline{underline}, respectively.}
\label{tab:robocasa}
\begin{tabular}{lc}
\toprule
\textbf{Method} & \textbf{Avg. Success (\%)} \\
\midrule
UVA~\cite{li2025unified} & 50.0 \\
UWM~\cite{zhu2025unified} & 60.8 \\
$\pi_{0.5}$~\cite{intelligence2025pi_05} & 62.1 \\
$\pi_{0}$-FAST~\cite{pertsch2025fast} & 63.6 \\
GR00T N1.6~\cite{gr00tn1_2025} & 66.2 \\
Cosmos Policy~\cite{kim2026cosmos} & 67.1 \\
RLDX-1~\cite{kim2026rldx} & 70.6 \\
World2Act~\cite{vuong2026world2act} & \underline{72.6} \\
\midrule
\textbf{Xiaomi-Robotics-1 (Ours)} & \textbf{74.5} \\
\bottomrule
\end{tabular}
\end{table}
\begin{table}[!t]
\centering
\small
\setlength{\tabcolsep}{6pt}
\caption{\textbf{Results on the RoboCasa365 benchmark.} We report task success rates (\%). The best and second-best results are highlighted in \textbf{bold} and \underline{underline}, respectively.}
\label{tab:robocasa365}

\begin{tabular}{
l
S[table-format=2.1]
S[table-format=2.1]
S[table-format=2.1]
S[table-format=2.1]
}
\toprule
\textbf{Method}
& {\textbf{Average}}
& {\textbf{Atomic}}
& {\textbf{Comp.-Seen}}
& {\textbf{Comp.-Unseen}} \\
\midrule

Diffusion Policy~\cite{chi2024diffusionpolicy}
& 6.1 & 15.7 & 0.2 & 1.3 \\

$\pi_{0.5}$~\cite{intelligence2025pi_05}
& 16.9 & 39.6 & 7.1 & 1.2 \\

GigaWorld-Policy 0.1~\cite{ye2026gigaworld}
& 20.7 & 44.4 & 11.8 & 2.9 \\

GR00T-N1.6~\cite{gr00tn1_2025}
& 21.9 & 51.1 & 9.4 & 1.7 \\

WorldDreamer~\cite{wang2023world}
& 35.3 & 66.3 & 26.7 & 9.0 \\

Qwen-RobotManip~\cite{qwenrobotmanip}
& 35.9 & 68.6 & 20.1 & \underline{14.9} \\

RLDX-1~\cite{kim2026rldx}
& 36.0 & 67.6 & 27.9 & 8.5 \\

ABot-M0.5~\cite{chen2026abotm05}
& 40.4 & 75.9 & 38.3 & 2.7 \\

ABot-M0.6~\cite{chen2026abotm05}
& \underline{46.6}
& \underline{79.4}
& \underline{48.3}
& 7.9 \\
\midrule

\textbf{Xiaomi-Robotics-1 (Ours)}
& \bfseries \textbf{57.4}
& \bfseries \textbf{80.2}
& \bfseries \textbf{57.1}
& \bfseries \textbf{32.1} \\
\bottomrule

\end{tabular}
\end{table}

\Ours{} achieves state-of-the-art results across all the four challenging benchmarks.
It achieves an average success rate of 74.5\% in RoboCasa.
In RoboCasa365, it significantly outperforms the previous best methods by 10.8 percentage points, obtaining an average success rate of 57.4\%.
Specifically, it delivers the largest improvement on the most challenging split of \textit{Composite-Unseen}, showcasing powerful generalization capabilities to task compositions that were unseen during training.
In VLABench, \Ours{} achieves the highest average success rate and progress score, while maintaining a competitive intention score.
This highlights its strong robustness to semantic and distribution shifts as well as superior long-horizon reasoning and language-conditioned manipulation capabilities.
Notably, under cross-category and texture shifts, \Ours{} surpasses the strongest baseline by 6.0 and 15.2 percentage points in terms of success rate, respectively, demonstrating effective generalization to unseen objects and high resilience to visual appearance and background perturbations.
In RoboDojo, \Ours{} significantly outperforms existing baseline methods, achieving absolute improvements of 7\% and 5.13\% in terms of average score and average success rate, respectively, over the second-best method.
It ranks first across four out of five dimensions, indicating the model’s outstanding overall performance across diverse task types.
Notably, we do not incorporate history observation in the evaluation of the RoboDojo benchmark. 
Consequently, it scores lower on the memory dimension than Hy-Embodied-0.5-VLA~\cite{zhang2026hy} which explicitly models memory, yet it still significantly outperforms all other models.

\begin{table*}[t]
\centering
\small
\setlength{\tabcolsep}{2.8pt}
\renewcommand{\arraystretch}{1.12}

\caption{\textbf{Results on the VLABench Benchmark.}
SR, PS, and IS denote success rate, progress score, and intention score, respectively.
The best and second-best results are highlighted in \textbf{bold} and \underline{underline}, respectively.
}
\label{tab:vlabench}

\begin{tabular}{lccccccccc}
\toprule

\multirow{2}{*}{\textbf{Method}}
& \multicolumn{3}{c}{\textbf{In-dist.}}
& \multicolumn{3}{c}{\textbf{Cross Category}}
& \multicolumn{3}{c}{\textbf{Commonsense}}
\\

\cmidrule(lr){2-4}
\cmidrule(lr){5-7}
\cmidrule(lr){8-10}

& SR $\uparrow$ & PS $\uparrow$ & IS $\uparrow$
& SR $\uparrow$ & PS $\uparrow$ & IS $\uparrow$
& SR $\uparrow$ & PS $\uparrow$ & IS $\uparrow$
\\
\midrule

$\pi_0$-FAST~\cite{pertsch2025fast}
& 56.2 & 66.8 & 72.4
& 31.0 & 38.2 & 47.8
& 38.0 & 48.6 & 56.8
\\

X-VLA~\cite{zheng2025x}
& -- & 67.8 & --
& -- & 25.1 & --
& -- & 48.2 & --
\\

ACoT-VLA~\cite{zhong2026acot}
& -- & 66.1 & \underline{79.8}
& -- & 38.9 & 54.1
& -- & 37.8 & 52.3
\\

$\pi_{0.5}$~\cite{intelligence2025pi_05}
& 65.4 & 77.8 & 80.4
& 38.2 & 49.7 & 52.0
& 43.9 & \underline{57.3} & \textbf{60.0}
\\

ERVLA~\cite{sun2026revisiting}
& \underline{69.7} & \underline{81.1} & \textbf{84.2}
& \underline{47.0} & \underline{61.0} & \underline{66.4}
& \underline{44.0} & 55.0 & 57.2
\\

\midrule

\textbf{Xiaomi-Robotics-1 (Ours)}
& \textbf{75.6} & \textbf{85.0} & \underline{79.8}
& \textbf{53.0} & \textbf{66.6} & \textbf{66.4}
& \textbf{48.4} & \textbf{58.3} & \underline{58.2}
\\

\bottomrule
\end{tabular}

\vspace{3mm}

\begin{tabular}{lccccccccc}
\toprule

\multirow{2}{*}{\textbf{Method}}
& \multicolumn{3}{c}{\textbf{Instruction}}
& \multicolumn{3}{c}{\textbf{Texture}}
& \multicolumn{3}{c}{\textbf{Avg.}}
\\

\cmidrule(lr){2-4}
\cmidrule(lr){5-7}
\cmidrule(lr){8-10}

& SR $\uparrow$ & PS $\uparrow$ & IS $\uparrow$
& SR $\uparrow$ & PS $\uparrow$ & IS $\uparrow$
& SR $\uparrow$ & PS $\uparrow$ & IS $\uparrow$
\\
\midrule

$\pi_0$-FAST~\cite{pertsch2025fast}
& 35.0 & 45.0 & 59.4
& 39.0 & 49.0 & 56.8
& 39.8 & 49.5 & 58.6
\\

X-VLA~\cite{zheng2025x}
& -- & 63.1 & --
& -- & -- & --
& -- & 51.1 & --
\\

ACoT-VLA~\cite{zhong2026acot}
& -- & 39.6 & 56.8
& -- & 54.6 & 74.6
& -- & 47.4 & 63.5
\\

$\pi_{0.5}$~\cite{intelligence2025pi_05}
& 48.2 & 64.2 & 67.0
& 44.9 & \underline{62.3} & 65.0
& 48.1 & 62.3 & 64.9
\\

ERVLA~\cite{sun2026revisiting}
& \textbf{58.0} & \textbf{70.2} & \textbf{73.8}
& \underline{47.4} & \underline{62.3} & \underline{70.6}
& \underline{53.2} & \underline{65.9} & \textbf{70.4}
\\

\midrule

\textbf{Xiaomi-Robotics-1 (Ours)}
& \underline{55.8} & \underline{66.8} & \underline{70.2}
& \textbf{62.6} & \textbf{74.9} & \textbf{74.8}
& \textbf{59.1} & \textbf{70.3} & \underline{69.9}
\\

\bottomrule
\end{tabular}

\end{table*}

\newcommand{\result}[2]{#1 \,/\, #2\%}
\newcommand{\bestresult}[2]{\textbf{#1} \,/\, \textbf{#2}\%}
\newcommand{\secondresult}[2]{\underline{#1} \,/\, \underline{#2}\%}

\begin{table*}[t]
\centering
\small
\setlength{\tabcolsep}{3.0pt}
\renewcommand{\arraystretch}{1.08}

\caption{
\textbf{Results on the RoboDojo Simulation Benchmark.}
Each entry reports score / success rate (\%).
The best and second-best results are highlighted in \textbf{bold} and \underline{underline}, respectively.
}
\label{tab:robodojo}

\begin{tabular*}{0.82\textwidth}{
    @{\extracolsep{\fill}}
    lccc
    @{}
}
\toprule

\textbf{Method}
& \textbf{Generalization}
& \textbf{Precision}
& \textbf{Long-Horizon}
\\
\midrule

GalaxeaVLA (G0)~\citep{jiang2025galaxea}
& \result{4.53}{2.83}
& \result{8.10}{3.83}
& \result{12.60}{5.58}
\\

GigaWorld-Policy~\citep{ye2026gigaworld}
& \result{5.34}{2.89}
& \result{6.15}{1.83}
& \result{15.51}{8.92}
\\

StarVLA-$\alpha$~\citep{ye2026starvla}
& \result{3.93}{2.33}
& \result{9.90}{4.33}
& \result{14.15}{6.50}
\\

Xiaomi-Robotics-0~\citep{cai2026xiaomi}
& \result{7.43}{5.56}
& \result{8.42}{4.58}
& \result{13.51}{6.92}
\\

X-WAM~\citep{guo2026xwam}
& \result{7.39}{3.33}
& \result{6.72}{1.83}
& \result{17.47}{9.08}
\\

X-VLA~\citep{zheng2025x}
& \result{10.48}{6.78}
& \secondresult{18.32}{12.00}
& \result{16.53}{9.75}
\\

$\pi_{0.5}$~\citep{intelligence2025pi_05}
& \result{13.37}{8.17}
& \result{12.40}{5.50}
& \result{23.54}{14.67}
\\

Spatial Forcing~\citep{spatialforcing2025}
& \secondresult{14.12}{9.33}
& \result{17.33}{10.58}
& \result{23.26}{14.58}
\\

Hy-Embodied-0.5-VLA~\citep{zhang2026hy}
& \result{11.77}{8.39}
& \result{13.81}{8.00}
& \secondresult{25.74}{14.92}
\\

\midrule

\textbf{Xiaomi-Robotics-1 (Ours)}
& \bestresult{23.55}{17.00}
& \bestresult{26.69}{18.83}
& \bestresult{38.39}{23.67}
\\

\bottomrule
\end{tabular*}

\vspace{4mm}

\begin{tabular*}{0.82\textwidth}{
    @{\extracolsep{\fill}}
    lccc
    @{}
}
\toprule

\textbf{Method}
& \textbf{Memory}
& \textbf{Open}
& \textbf{Average}
\\
\midrule

GalaxeaVLA (G0)~\citep{jiang2025galaxea}
& \result{3.17}{1.89}
& \result{0.70}{0.67}
& \result{5.82}{2.96}
\\

GigaWorld-Policy~\citep{ye2026gigaworld}
& \result{3.46}{2.22}
& \result{0.54}{0.50}
& \result{6.20}{3.27}
\\

StarVLA-$\alpha$~\citep{ye2026starvla}
& \result{3.34}{2.44}
& \result{0.68}{0.58}
& \result{6.40}{3.24}
\\

Xiaomi-Robotics-0~\citep{cai2026xiaomi}
& \result{5.07}{3.67}
& \result{0.22}{0.17}
& \result{6.93}{4.18}
\\

X-WAM~\citep{guo2026xwam}
& \result{6.32}{4.67}
& \result{0.57}{0.25}
& \result{7.69}{3.83}
\\

X-VLA~\citep{zheng2025x}
& \result{4.76}{3.56}
& \result{0.55}{0.50}
& \result{10.13}{6.52}
\\

$\pi_{0.5}$~\citep{intelligence2025pi_05}
& \result{5.78}{4.56}
& \secondresult{1.98}{1.67}
& \result{11.41}{6.91}
\\

Spatial Forcing~\citep{spatialforcing2025}
& \result{5.43}{4.11}
& \result{1.78}{1.58}
& \result{12.38}{8.04}
\\

Hy-Embodied-0.5-VLA~\citep{zhang2026hy}
& \bestresult{13.37}{12.11}
& \result{0.65}{0.58}
& \secondresult{13.07}{8.80}
\\

\midrule

\textbf{Xiaomi-Robotics-1 (Ours)}
& \secondresult{7.81}{6.56}
& \bestresult{3.94}{3.58}
& \bestresult{20.07}{13.93}
\\

\bottomrule
\end{tabular*}

\end{table*}

\section{Related Work}
\label{related_work}

\paragraph{Scaling for Robot Learning} 
Research on the scaling laws of large language models (LLMs) demonstrates that performance improves predictably when data, compute, and model capacity are scaled in tandem~\cite{kaplan2020scaling, hoffmann2022training}.
Recent advances in large language models~\cite{brown2020language, touvron2023llama} and multi-modal foundation models~\cite{achiam2023gpt, team2024gemini, bai2025qwen3, agarwal2026cosmos} further demonstrate substantial capability gains driven by scaling of data and model size.
Motivated by these advancements, robot learning has increasingly embraced this scaling paradigm, enlarging both data and model sizes in pursuit of foundational generalist policies~\cite{brohan2022rt, zitkovich2023rt2, black2024pi_0, intelligence2025pi_05, pertsch2025fast, liu2025towards, gr00tn1_2025, kim2025fine, cai2026xiaomi,generalist2025gen0,generalist2026gen1,genesis2026gene265}.
Scaling robot learning, however, differs fundamentally from scaling models trained on web-scale data.
Collecting real-robot trajectories requires costly and labor-intensive teleoperation, and the resulting data are often confined to a narrow slice of environments and tasks, severely constraining the data diversity.
To alleviate this data bottleneck, recent work leverages portable UMI devices~\cite{chi2024universal}, enabling the collection of real-world manipulation trajectories without requiring physical robot embodiments~\cite{chi2024universal, liu2025vitamin, xu2025dexumi,zhaxizhuoma2025fastumi, generalist2025gen0, generalist2026gen1}.
This lowers the barrier to data collection by facilitating in-the-wild collection across a highly diverse spectrum of unconstrained, real-world environments.
Complementarily, another line of research leverages egocentric human manipulation trajectories to capture diverse behaviors across various tasks, objects, and environments, and subsequently converts them to robot data via representation alignment or motion retargeting~\cite{luo2025being, li2025scalable, chen2026wh0}.
In this work, we leverage over 100k hours of real-world manipulation trajectories collected from UMI devices.
By varying data scale and model size, we investigate the scaling behavior of our model during pre-training and whether scaling translates to real-robot performance after post-training.

\paragraph{Robot Foundation Models}
Recently, robot foundation models have emerged as a new paradigm for generalizable robot learning.
By leveraging large-scale datasets, these models enable robust generalization across diverse environments and facilitate efficient adaptation to novel downstream tasks.
World-action models (WAMs)~\cite{ye2026world,yuan2026fast,kim2026cosmos,pai2025mimic,guo2026xwam,li2026multi,li2026causal,zhang2026native,team2026motubrain,yang2026memorywam,ma2026dit4dit} and vision-language-action (VLA) models~\cite{black2024pi_0,intelligence2025pi_05,cheang2024gr,cheang2025gr,liu2025towards,yu2026wall,kim2024openvla,zheng2025x,qu2025spatialvla,li2026bridgevla,cai2026xiaomi} represent two popular paradigms of robot foundation models.
WAMs are generally built upon pre-trained video models.
They explicitly model future observations or environment dynamics and use these predictions to facilitate action generation. 
Early approaches typically generate visual plans from language-specified goals and subsequently translate them into executable robot actions~\cite{du2023learning, ko2024learning, zhou2024robodreamer, liang2024dreamitate,li2025gr}.
More recent work extends this paradigm to several directions, including jointly modeling future prediction and action generation~\cite{hu2024video, li2025unified, zhu2025unified, li2026causal, kim2026cosmos}, incorporating explicit 3D geometric structure~\cite{zhen2025tesseract, guo2026xwam, zhao2026rynnworld, li2026multi}, and training on large-scale heterogeneous video-action data to improve zero-shot generalization and cross-embodiment transfer~\cite{zhang2026native, ye2026world}.
In contrast, VLA models leverage pre-trained vision-language models as backbone, aiming to harness their rich, general semantic knowledge to facilitate robust action prediction~\cite{zitkovich2023rt2, kim2024openvla, black2024pi_0, cheang2024gr, gr00tn1_2025, cheang2025gr, cai2026xiaomi, liu2025rdt, intelligence2025pi_05, intelligence2025pi_06star, intelligence2026pi_07}.
Recent advances improve VLA models along three main axes. 
First, reasoning-oriented methods introduce intermediate representations (\textit{e.g.}, embodied reasoning tokens~\cite{team2025gemini, lee2025molmoact, fang2026molmoact2, chen2025training, zawalski2024robotic} and visual chains of thought~\cite{zhao2025cot, zheng2025tracevla}) to support semantic, spatial, and temporal reasoning. 
Second, advances in action representation address the limitations of naive action discretization through learned tokenizers and flow matching~\cite{pertsch2025fast, black2024pi_0, galaxea2026g05, yu2026wall, liu2026rdt2}.
Third, large-scale pre-training across heterogeneous datasets with different robot embodiments enables strong cross-embodiment transfer and robust downstream performance~\cite{o2024open, team2024octo, kim2024openvla, gr00tn1_2025, cai2026xiaomi, yu2026wall}.
Our work follows the VLA paradigm but focuses on a complementary question: the scaling behavior of robot foundation models.
To support this investigation, we develop a scalable infrastructure to curate massive trajectory datasets with precise language annotations.
Leveraging this data for large-scale training, we conduct extensive experiments to systematically explore the scaling properties of foundational VLA models.

\section{Conclusions}
In this work, we present \Ours{}, a foundational vision-language-action (VLA) model that is able to follow instructions to perform a wide range of mobile manipulation tasks out-of-the-box in unseen environments, and efficiently adapt to novel challenging tasks with a minimal amount of data.
During pre-training, we leverage over 100,000 hours of real-world manipulation trajectories, endowing the model with broad and generalizable manipulation capabilities.
To scale training effectively, we propose an auto-labeling pipeline that annotates the large-scale dataset with detailed descriptions of scene state transitions as language prompts.
In the post-training phase, we align these strong capabilities acquired during pre-training with robot embodiments and imperative instruction prompts using a cross-embodiment dataset.
Extensive experiments demonstrate that the performance of \Ours{} consistently improves with increasing data scale and model size during pre-training.
More importantly, the scaling property directly translates to out-of-the-box performance in unseen environments after post-training.
\Ours{} can also serve as a powerful robot foundation model that is able to adapt to novel challenging real-robot tasks with minimal data.
In addition, it achieves strong state-of-the-art results on four challenging simulation benchmarks.
We hope this work can serve as a foundation for future exploration of scalable robot policies that can be deployed out-of-the-box in the real world.

\clearpage

\section*{Contributions}
\label{contributions}
\setlength{\parskip}{5pt} 
\setlength{\itemsep}{0pt} 
Authors are listed in alphabetical order. 
\vspace{\baselineskip}

\begin{multicols}{-2}
\textbf{\fontsize{10pt}{14pt}\selectfont Core Contributors}
\begin{itemize}
    \item Jun Guo
    \item Piaopiao Jin
    \item Jason Li
    \item Peiyan Li
    \item Yingyan Li
    \item Futeng Liu
    \item Wanli Peng
    \item Optimus Qin
    \item Yifei Su
    \item Nan Sun
    \item Qiao Sun
    \item Runze Suo
    \item Heyun Wang
    \item Yunhong Wang
    \item Rujie Wu
    \item Caoyu Xia
    \item Lina Zhang
    \item Jack Zhao
\end{itemize}

\columnbreak 

\textbf{\fontsize{10pt}{14pt}\selectfont Contributors}
\begin{itemize}
    \item Guoliang Chen
    \item Wenlong Chen
    \item Xinze He
    \item Bin Li
    \item Qing Li
    \item Zhuorong Li
    \item Heng Qu
    \item Wenxuan Song
    \item Diyun Xiang
    \item Yifan Xie
    \item Peiran Xu
    \item Hangjun Ye
    \item Wen Ye
    \item Han Zhao
    \item Quanyun Zhou
\end{itemize}

\end{multicols}

\section*{Acknowledgment}
\label{Acknowledgment}

We express our sincere appreciation to the broader team for their tremendous support, including those not listed above: Li Jiang, Xiaohan Yu, Meichen Mu, Xiaoke Xilinjueluo, Qingyi Li, Qi Liu, Yayun Liu, Jun Xia, Feng Qiu, Donghao Wang, Yan Hou, Dong Wang, Liangliang He, Jiaxin Liu, Kang Zhou, Rui Cai, Shuoxue Bi, Yingchao Zhou, Kun Ma, Yiwei Zhou and Dongsheng Li.

\clearpage

\bibliographystyle{plainnat}
\bibliography{main}

\begin{thebibliography}{97}
\providecommand{\natexlab}[1]{#1}
\providecommand{\url}[1]{\texttt{#1}}
\expandafter\ifx\csname urlstyle\endcsname\relax
  \providecommand{\doi}[1]{doi: #1}\else
  \providecommand{\doi}{doi: \begingroup \urlstyle{rm}\Url}\fi

\bibitem[Achiam et~al.(2023)Achiam, Adler, Agarwal, Ahmad, Akkaya, Aleman, Almeida, Altenschmidt, Altman, Anadkat, et~al.]{achiam2023gpt}
Josh Achiam, Steven Adler, Sandhini Agarwal, Lama Ahmad, Ilge Akkaya, Florencia~Leoni Aleman, Diogo Almeida, Janko Altenschmidt, Sam Altman, Shyamal Anadkat, et~al.
\newblock Gpt-4 technical report.
\newblock \emph{arXiv preprint arXiv:2303.08774}, 2023.

\bibitem[Agarwal et~al.(2026)Agarwal, Ali, Allen, Antolini, Aubame, Azzolini, Bai, Bala, Balaji, Bapst, et~al.]{agarwal2026cosmos}
Niket Agarwal, Arslan Ali, Jon Allen, Martin Antolini, Adeline Aubame, Alisson Azzolini, Junjie Bai, Maciej Bala, Yogesh Balaji, Josh Bapst, et~al.
\newblock Cosmos 3: Omnimodal world models for physical ai.
\newblock \emph{arXiv preprint arXiv:2606.02800}, 2026.

\bibitem[Bai et~al.(2025)Bai, Cai, Chen, Chen, Chen, Cheng, Deng, Ding, Gao, Ge, et~al.]{bai2025qwen3}
Shuai Bai, Yuxuan Cai, Ruizhe Chen, Keqin Chen, Xionghui Chen, Zesen Cheng, Lianghao Deng, Wei Ding, Chang Gao, Chunjiang Ge, et~al.
\newblock Qwen3-vl technical report.
\newblock \emph{arXiv preprint arXiv:2511.21631}, 2025.

\bibitem[Black et~al.(2024)Black, Brown, Driess, Esmail, Equi, Finn, Fusai, Groom, Hausman, Ichter, et~al.]{black2024pi_0}
Kevin Black, Noah Brown, Danny Driess, Adnan Esmail, Michael Equi, Chelsea Finn, Niccolo Fusai, Lachy Groom, Karol Hausman, Brian Ichter, et~al.
\newblock {$\pi_{0}$}: A vision-language-action flow model for general robot control.
\newblock \emph{arXiv preprint arXiv:2410.24164}, 2024.

\bibitem[Black et~al.(2025)Black, Brown, Darpinian, Dhabalia, Driess, Esmail, Equi, Finn, Fusai, et~al.]{intelligence2025pi_05}
Kevin Black, Noah Brown, James Darpinian, Karan Dhabalia, Danny Driess, Adnan Esmail, Michael Equi, Chelsea Finn, Niccolo Fusai, et~al.
\newblock {$\pi_{0.5}$}: a vision-language-action model with open-world generalization.
\newblock \emph{arXiv preprint arXiv:2504.16054}, 2025.

\bibitem[Brohan et~al.(2022)Brohan, Brown, Carbajal, Chebotar, Dabis, Finn, Gopalakrishnan, Hausman, Herzog, Hsu, et~al.]{brohan2022rt}
Anthony Brohan, Noah Brown, Justice Carbajal, Yevgen Chebotar, Joseph Dabis, Chelsea Finn, Keerthana Gopalakrishnan, Karol Hausman, Alex Herzog, Jasmine Hsu, et~al.
\newblock Rt-1: Robotics transformer for real-world control at scale.
\newblock \emph{arXiv preprint arXiv:2212.06817}, 2022.

\bibitem[Brown et~al.(2020)Brown, Mann, Ryder, Subbiah, Kaplan, Dhariwal, Neelakantan, Shyam, Sastry, Askell, et~al.]{brown2020language}
Tom Brown, Benjamin Mann, Nick Ryder, Melanie Subbiah, Jared~D Kaplan, Prafulla Dhariwal, Arvind Neelakantan, Pranav Shyam, Girish Sastry, Amanda Askell, et~al.
\newblock Language models are few-shot learners.
\newblock \emph{Advances in neural information processing systems}, 33:\penalty0 1877--1901, 2020.

\bibitem[Cai et~al.(2026)Cai, Guo, He, Jin, Li, Lin, Liu, Liu, Ma, Ma, et~al.]{cai2026xiaomi}
Rui Cai, Jun Guo, Xinze He, Piaopiao Jin, Jie Li, Bingxuan Lin, Futeng Liu, Wei Liu, Fei Ma, Kun Ma, et~al.
\newblock Xiaomi-robotics-0: An open-sourced vision-language-action model with real-time execution.
\newblock \emph{arXiv preprint arXiv:2602.12684}, 2026.

\bibitem[Cheang et~al.(2024)Cheang, Chen, Jing, Kong, Li, Li, Liu, Wu, Xu, Yang, et~al.]{cheang2024gr}
Chi-Lam Cheang, Guangzeng Chen, Ya~Jing, Tao Kong, Hang Li, Yifeng Li, Yuxiao Liu, Hongtao Wu, Jiafeng Xu, Yichu Yang, et~al.
\newblock Gr-2: A generative video-language-action model with web-scale knowledge for robot manipulation.
\newblock \emph{arXiv preprint arXiv:2410.06158}, 2024.

\bibitem[Cheang et~al.(2025)Cheang, Chen, Cui, Hu, Huang, Kong, Li, Li, Liu, Ma, et~al.]{cheang2025gr}
Chilam Cheang, Sijin Chen, Zhongren Cui, Yingdong Hu, Liqun Huang, Tao Kong, Hang Li, Yifeng Li, Yuxiao Liu, Xiao Ma, et~al.
\newblock Gr-3 technical report.
\newblock \emph{arXiv preprint arXiv:2507.15493}, 2025.

\bibitem[Chen et~al.(2026{\natexlab{a}})Chen, Yang, Tang, Huo, Lin, Wu, Liu, Chen, Zheng, Yuan, Li, Wang, Qi, Hu, Mei, Xuan, Yang, Zhu, Xu, Ma, and Chang]{chen2026abotm05}
Ronghan Chen, Yandan Yang, Zuojin Tang, Dongjie Huo, Tong Lin, Haoning Wu, Haoyun Liu, Yuzhi Chen, Lulu Zheng, Botai Yuan, Tianlun Li, Mingxin Wang, Dekang Qi, Bin Hu, Wei Mei, Yuze Xuan, Haolong Yang, Yanqing Zhu, Mu~Xu, Zhiheng Ma, and Xinyuan Chang.
\newblock Abot-m0.5: Unified mobility-and-manipulation world action model.
\newblock \emph{arXiv preprint arXiv:2607.00678}, 2026{\natexlab{a}}.

\bibitem[Chen et~al.(2026{\natexlab{b}})Chen, Chen, Li, Tang, Su, Lu, Wan, Chen, Liu, Yan, Su, Dou, Wang, Zhang, Liu, Qin, Liang, Wu, Lin, Lin, Wang, He, Wu, Wu, Zhou, Lei, Yu, Ji, Jin, Lin, Li, Xiong, Xu, Li, Chai, Xie, Wang, Mu, Dong, Matusik, Ding, Ding, Luo, and Tomizuka]{chen2026robodojounifiedsimandrealbenchmark}
Tianxing Chen, Yue Chen, Zixuan Li, Junyuan Tang, Kailun Su, Haoran Lu, Weijie Wan, Baijun Chen, Songling Liu, Haowen Yan, Honghao Su, Zhiyang Dou, Kaixuan Wang, Dandan Zhang, Yunze Liu, Yan Qin, Qiwei Liang, Qiwei Wu, Zijian Lin, Wenwei Lin, Yuran Wang, Minghua He, Tianshu Wu, Ruihai Wu, Jingquan Zhou, Kai-Chong Lei, Haibao Yu, Yuanfeng Ji, Weiyang Jin, Guanyu Lin, Xiaofan Li, Qi~Xiong, Renjing Xu, Zhongyu Li, Wenhao Chai, Enze Xie, Ziwei Wang, Yao Mu, Hao Dong, Wojciech Matusik, Mingyu Ding, Wenbo Ding, Ping Luo, and Masayoshi Tomizuka.
\newblock Robodojo: A unified sim-and-real benchmark for comprehensive evaluation of generalist robot manipulation policies, 2026{\natexlab{b}}.
\newblock URL \url{https://arxiv.org/abs/2607.04434}.

\bibitem[Chen et~al.(2025)Chen, Belkhale, Mirchandani, Mees, Driess, Pertsch, and Levine]{chen2025training}
William Chen, Suneel Belkhale, Suvir Mirchandani, Oier Mees, Danny Driess, Karl Pertsch, and Sergey Levine.
\newblock Training strategies for efficient embodied reasoning.
\newblock \emph{arXiv preprint arXiv:2505.08243}, 2025.

\bibitem[Chen et~al.(2022)Chen, Wang, Changpinyo, Piergiovanni, Padlewski, Salz, Goodman, Grycner, Mustafa, Beyer, et~al.]{chen2022pali}
Xi~Chen, Xiao Wang, Soravit Changpinyo, Anthony~J Piergiovanni, Piotr Padlewski, Daniel Salz, Sebastian Goodman, Adam Grycner, Basil Mustafa, Lucas Beyer, et~al.
\newblock Pali: A jointly-scaled multilingual language-image model.
\newblock \emph{arXiv preprint arXiv:2209.06794}, 2022.

\bibitem[Chen et~al.(2026{\natexlab{c}})Chen, Chen, Wang, Li, Huo, Shi, and Gao]{chen2026wh0}
Yangtao Chen, Zixuan Chen, Peiyang Wang, Yong-Lu Li, Jing Huo, Jieqi Shi, and Yang Gao.
\newblock Wh0: Generative world models as scalable sources of egocentric human hand manipulation data.
\newblock \emph{arXiv preprint arXiv:2606.22136}, 2026{\natexlab{c}}.

\bibitem[Chi et~al.(2024{\natexlab{a}})Chi, Xu, Feng, Cousineau, Du, Burchfiel, Tedrake, and Song]{chi2024diffusionpolicy}
Cheng Chi, Zhenjia Xu, Siyuan Feng, Eric Cousineau, Yilun Du, Benjamin Burchfiel, Russ Tedrake, and Shuran Song.
\newblock Diffusion policy: Visuomotor policy learning via action diffusion.
\newblock \emph{The International Journal of Robotics Research}, 2024{\natexlab{a}}.

\bibitem[Chi et~al.(2024{\natexlab{b}})Chi, Xu, Pan, Cousineau, Burchfiel, Feng, Tedrake, and Song]{chi2024universal}
Cheng Chi, Zhenjia Xu, Chuer Pan, Eric Cousineau, Benjamin Burchfiel, Siyuan Feng, Russ Tedrake, and Shuran Song.
\newblock Universal manipulation interface: In-the-wild robot teaching without in-the-wild robots.
\newblock \emph{arXiv preprint arXiv:2402.10329}, 2024{\natexlab{b}}.

\bibitem[Du et~al.(2023)Du, Yang, Dai, Dai, Nachum, Tenenbaum, Schuurmans, and Abbeel]{du2023learning}
Yilun Du, Sherry Yang, Bo~Dai, Hanjun Dai, Ofir Nachum, Josh Tenenbaum, Dale Schuurmans, and Pieter Abbeel.
\newblock Learning universal policies via text-guided video generation.
\newblock \emph{Advances in neural information processing systems}, 36:\penalty0 9156--9172, 2023.

\bibitem[Fang et~al.(2026)Fang, Duan, Clay, Wang, Liu, Huang, Fan, Tsai, Chen, Wang, et~al.]{fang2026molmoact2}
Haoquan Fang, Jiafei Duan, Donovan Clay, Sam Wang, Shuo Liu, Weikai Huang, Xiang Fan, Wei-Chuan Tsai, Shirui Chen, Yi~Ru Wang, et~al.
\newblock Molmoact2: Action reasoning models for real-world deployment.
\newblock \emph{arXiv preprint arXiv:2605.02881}, 2026.

\bibitem[{Galaxea Team}(2026)]{galaxea2026g05}
{Galaxea Team}.
\newblock Galaxea g0.5 technical report.
\newblock 2026.
\newblock URL \url{https://opengalaxea.github.io/G05/}.

\bibitem[Guo et~al.(2026)Guo, Li, Li, Chen, Sun, Su, Wang, Zhang, Li, and Liu]{guo2026xwam}
Jun Guo, Qiwei Li, Peiyan Li, Zilong Chen, Nan Sun, Yifei Su, Heyun Wang, Yuan Zhang, Xinghang Li, and Huaping Liu.
\newblock Unified 4d world action modeling from video priors with asynchronous denoising.
\newblock \emph{arXiv preprint arXiv:2604.26694}, 2026.

\bibitem[Hoffmann et~al.(2022)Hoffmann, Borgeaud, Mensch, Buchatskaya, Cai, Rutherford, Casas, Hendricks, Welbl, Clark, et~al.]{hoffmann2022training}
Jordan Hoffmann, Sebastian Borgeaud, Arthur Mensch, Elena Buchatskaya, Trevor Cai, Eliza Rutherford, DDL Casas, Lisa~Anne Hendricks, Johannes Welbl, Aidan Clark, et~al.
\newblock Training compute-optimal large language models.
\newblock \emph{arXiv preprint arXiv:2203.15556}, 10, 2022.

\bibitem[Hu et~al.(2024)Hu, Guo, Wang, Chen, Wang, Zhang, Sreenath, Lu, and Chen]{hu2024video}
Yucheng Hu, Yanjiang Guo, Pengchao Wang, Xiaoyu Chen, Yen-Jen Wang, Jianke Zhang, Koushil Sreenath, Chaochao Lu, and Jianyu Chen.
\newblock Video prediction policy: A generalist robot policy with predictive visual representations.
\newblock \emph{arXiv preprint arXiv:2412.14803}, 2024.

\bibitem[Intelligence et~al.(2025)Intelligence, Amin, Aniceto, Balakrishna, Black, Conley, Connors, Darpinian, Dhabalia, DiCarlo, et~al.]{intelligence2025pi_06star}
Physical Intelligence, Ali Amin, Raichelle Aniceto, Ashwin Balakrishna, Kevin Black, Ken Conley, Grace Connors, James Darpinian, Karan Dhabalia, Jared DiCarlo, et~al.
\newblock {$\pi^{*}_{0.6}$}: a vla that learns from experience.
\newblock \emph{arXiv preprint arXiv:2511.14759}, 2025.

\bibitem[Intelligence et~al.(2026)Intelligence, Ai, Amin, Aniceto, Balakrishna, Balke, Black, Bokinsky, Cao, Charbonnier, et~al.]{intelligence2026pi_07}
Physical Intelligence, Bo~Ai, Ali Amin, Raichelle Aniceto, Ashwin Balakrishna, Greg Balke, Kevin Black, George Bokinsky, Shihao Cao, Thomas Charbonnier, et~al.
\newblock {$\pi_{0.7}$}: a steerable generalist robotic foundation model with emergent capabilities.
\newblock \emph{arXiv preprint arXiv:2604.15483}, 2026.

\bibitem[Jiang et~al.(2025)Jiang, Yuan, Liu, Lu, Cui, Liu, Cheng, Gao, Xu, and Zhao]{jiang2025galaxea}
Tao Jiang, Tianyuan Yuan, Yicheng Liu, Chenhao Lu, Jianning Cui, Xiao Liu, Shuiqi Cheng, Jiyang Gao, Huazhe Xu, and Hang Zhao.
\newblock Galaxea open-world dataset and g0 dual-system vla model.
\newblock \emph{arXiv preprint arXiv:2509.00576}, 2025.

\bibitem[Kaplan et~al.(2020)Kaplan, McCandlish, Henighan, Brown, Chess, Child, Gray, Radford, Wu, and Amodei]{kaplan2020scaling}
Jared Kaplan, Sam McCandlish, Tom Henighan, Tom~B Brown, Benjamin Chess, Rewon Child, Scott Gray, Alec Radford, Jeffrey Wu, and Dario Amodei.
\newblock Scaling laws for neural language models.
\newblock \emph{arXiv preprint arXiv:2001.08361}, 2020.

\bibitem[Khazatsky et~al.(2024)Khazatsky, Pertsch, Nair, Balakrishna, Dasari, Karamcheti, Nasiriany, Srirama, Chen, Ellis, et~al.]{khazatsky2024droid}
Alexander Khazatsky, Karl Pertsch, Suraj Nair, Ashwin Balakrishna, Sudeep Dasari, Siddharth Karamcheti, Soroush Nasiriany, Mohan~Kumar Srirama, Lawrence~Yunliang Chen, Kirsty Ellis, et~al.
\newblock Droid: A large-scale in-the-wild robot manipulation dataset.
\newblock \emph{arXiv preprint arXiv:2403.12945}, 2024.

\bibitem[Kim et~al.(2026{\natexlab{a}})Kim, Jang, Koo, Jang, Kim, Kim, Yoon, Jang, Choi, Han, et~al.]{kim2026rldx}
Dongyoung Kim, Huiwon Jang, Myungkyu Koo, Suhyeok Jang, Taeyoung Kim, Beomjun Kim, Byungjun Yoon, Changsung Jang, Daewon Choi, Dongsu Han, et~al.
\newblock Rldx-1 technical report.
\newblock \emph{arXiv preprint arXiv:2605.03269}, 2026{\natexlab{a}}.

\bibitem[Kim et~al.(2024)Kim, Pertsch, Karamcheti, Xiao, Balakrishna, Nair, Rafailov, Foster, Lam, Sanketi, et~al.]{kim2024openvla}
Moo~Jin Kim, Karl Pertsch, Siddharth Karamcheti, Ted Xiao, Ashwin Balakrishna, Suraj Nair, Rafael Rafailov, Ethan Foster, Grace Lam, Pannag Sanketi, et~al.
\newblock Openvla: An open-source vision-language-action model.
\newblock \emph{arXiv preprint arXiv:2406.09246}, 2024.

\bibitem[Kim et~al.(2025)Kim, Finn, and Liang]{kim2025fine}
Moo~Jin Kim, Chelsea Finn, and Percy Liang.
\newblock Fine-tuning vision-language-action models: Optimizing speed and success.
\newblock \emph{arXiv preprint arXiv:2502.19645}, 2025.

\bibitem[Kim et~al.(2026{\natexlab{b}})Kim, Gao, Lin, Lin, Ge, Lam, Liang, Song, Liu, Finn, et~al.]{kim2026cosmos}
Moo~Jin Kim, Yihuai Gao, Tsung-Yi Lin, Yen-Chen Lin, Yunhao Ge, Grace Lam, Percy Liang, Shuran Song, Ming-Yu Liu, Chelsea Finn, et~al.
\newblock Cosmos policy: Fine-tuning video models for visuomotor control and planning.
\newblock \emph{arXiv preprint arXiv:2601.16163}, 2026{\natexlab{b}}.

\bibitem[Ko et~al.(2024)Ko, Mao, Du, Sun, and Tenenbaum]{ko2024learning}
Po-Chen Ko, Jiayuan Mao, Yilun Du, Shao-Hua Sun, and Joshua~B Tenenbaum.
\newblock Learning to act from actionless videos through dense correspondences.
\newblock In \emph{International Conference on Learning Representations}, volume 2024, pages 40938--40958, 2024.

\bibitem[Lee et~al.(2025)Lee, Duan, Fang, Deng, Liu, Li, Fang, Zhang, Wang, Lee, et~al.]{lee2025molmoact}
Jason Lee, Jiafei Duan, Haoquan Fang, Yuquan Deng, Shuo Liu, Boyang Li, Bohan Fang, Jieyu Zhang, Yi~Ru Wang, Sangho Lee, et~al.
\newblock Molmoact: Action reasoning models that can reason in space.
\newblock \emph{arXiv preprint arXiv:2508.07917}, 2025.

\bibitem[Li et~al.(2025{\natexlab{a}})Li, Song, Zhao, Wang, Ding, Wang, Zeng, and Li]{spatialforcing2025}
Fuhao Li, Wenxuan Song, Han Zhao, Jingbo Wang, Pengxiang Ding, Donglin Wang, Long Zeng, and Haoang Li.
\newblock Spatial forcing: Implicit spatial representation alignment for vision-language-action model.
\newblock \emph{arXiv preprint arXiv:2510.12276}, 2025{\natexlab{a}}.

\bibitem[Li et~al.(2026{\natexlab{a}})Li, Zhang, Luo, Yang, Wang, Han, Yu, Gao, Xue, Zhu, et~al.]{li2026causal}
Lin Li, Qihang Zhang, Yiming Luo, Shuai Yang, Ruilin Wang, Fei Han, Mingrui Yu, Zelin Gao, Nan Xue, Xing Zhu, et~al.
\newblock Causal world modeling for robot control.
\newblock \emph{arXiv preprint arXiv:2601.21998}, 2026{\natexlab{a}}.

\bibitem[Li et~al.(2025{\natexlab{b}})Li, Wu, Huang, Cheang, Wang, and Kong]{li2025gr}
Peiyan Li, Hongtao Wu, Yan Huang, Chilam Cheang, Liang Wang, and Tao Kong.
\newblock Gr-mg: Leveraging partially-annotated data via multi-modal goal-conditioned policy.
\newblock \emph{IEEE Robotics and Automation Letters}, 10\penalty0 (2):\penalty0 1912--1919, 2025{\natexlab{b}}.

\bibitem[Li et~al.(2026{\natexlab{b}})Li, Chen, Wu, Ma, Wu, Huang, Wang, Kong, and Tan]{li2026bridgevla}
Peiyan Li, Yixiang Chen, Hongtao Wu, Xiao Ma, Xiangnan Wu, Yan Huang, Liang Wang, Tao Kong, and Tieniu Tan.
\newblock Bridgevla: Input-output alignment for efficient 3d manipulation learning with vision-language models.
\newblock \emph{Advances in Neural Information Processing Systems}, 38:\penalty0 63635--63673, 2026{\natexlab{b}}.

\bibitem[Li et~al.(2026{\natexlab{c}})Li, Chen, Xu, Yang, Wu, Guo, Sun, Qian, Li, Xiao, et~al.]{li2026multi}
Peiyan Li, Yixiang Chen, Yuan Xu, Jiabing Yang, Xiangnan Wu, Jun Guo, Nan Sun, Long Qian, Xinghang Li, Xin Xiao, et~al.
\newblock Multi-view video diffusion policy: A 3d spatio-temporal-aware video action model.
\newblock \emph{arXiv preprint arXiv:2604.03181}, 2026{\natexlab{c}}.

\bibitem[Li et~al.(2025{\natexlab{c}})Li, Deng, Liang, Luo, Zhou, Yao, Zeng, Feng, Liang, Xu, et~al.]{li2025scalable}
Qixiu Li, Yu~Deng, Yaobo Liang, Lin Luo, Lei Zhou, Chengtang Yao, Lingqi Zeng, Zhiyuan Feng, Huizhi Liang, Sicheng Xu, et~al.
\newblock Scalable vision-language-action model pretraining for robotic manipulation with real-life human activity videos.
\newblock \emph{arXiv preprint arXiv:2510.21571}, 2025{\natexlab{c}}.

\bibitem[Li et~al.(2025{\natexlab{d}})Li, Gao, Sadigh, and Song]{li2025unified}
Shuang Li, Yihuai Gao, Dorsa Sadigh, and Shuran Song.
\newblock Unified video action model.
\newblock \emph{arXiv preprint arXiv:2503.00200}, 2025{\natexlab{d}}.

\bibitem[Li et~al.(2024)Li, Li, Liu, Wang, Liu, Kang, Ma, Kong, Zhang, and Liu]{liu2025towards}
Xinghang Li, Peiyan Li, Minghuan Liu, Dong Wang, Jirong Liu, Bingyi Kang, Xiao Ma, Tao Kong, Hanbo Zhang, and Huaping Liu.
\newblock Towards generalist robot policies: What matters in building vision-language-action models.
\newblock \emph{arXiv preprint arXiv:2412.14058}, 2024.

\bibitem[Liang et~al.(2024{\natexlab{a}})Liang, Liu, Ozguroglu, Sudhakar, Dave, Tokmakov, Song, and Vondrick]{liang2024dreamitate}
Junbang Liang, Ruoshi Liu, Ege Ozguroglu, Sruthi Sudhakar, Achal Dave, Pavel Tokmakov, Shuran Song, and Carl Vondrick.
\newblock Dreamitate: Real-world visuomotor policy learning via video generation.
\newblock \emph{arXiv preprint arXiv:2406.16862}, 2024{\natexlab{a}}.

\bibitem[Liang et~al.(2024{\natexlab{b}})Liang, Yu, Luo, Iyer, Dong, Zhou, Ghosh, Lewis, Yih, Zettlemoyer, et~al.]{liang2024mixture}
Weixin Liang, Lili Yu, Liang Luo, Srinivasan Iyer, Ning Dong, Chunting Zhou, Gargi Ghosh, Mike Lewis, Wen-tau Yih, Luke Zettlemoyer, et~al.
\newblock Mixture-of-transformers: A sparse and scalable architecture for multi-modal foundation models.
\newblock \emph{arXiv preprint arXiv:2411.04996}, 2024{\natexlab{b}}.

\bibitem[Liu et~al.(2024)Liu, Feng, Xue, Wang, Wu, Lu, Zhao, Deng, Zhang, Ruan, et~al.]{liu2024deepseek}
Aixin Liu, Bei Feng, Bing Xue, Bingxuan Wang, Bochao Wu, Chengda Lu, Chenggang Zhao, Chengqi Deng, Chenyu Zhang, Chong Ruan, et~al.
\newblock Deepseek-v3 technical report.
\newblock \emph{arXiv preprint arXiv:2412.19437}, 2024.

\bibitem[Liu et~al.(2025{\natexlab{a}})Liu, Li, Qin, Xu, Abbeel, and Chen]{liu2025vitamin}
Fangchen Liu, Chuanyu Li, Yihua Qin, Jing Xu, Pieter Abbeel, and Rui Chen.
\newblock Vitamin: Learning contact-rich tasks through robot-free visuo-tactile manipulation interface.
\newblock \emph{arXiv preprint arXiv:2504.06156}, 2025{\natexlab{a}}.

\bibitem[Liu et~al.(2025{\natexlab{b}})Liu, Wu, Li, Tan, Chen, Wang, Xu, Su, and Zhu]{liu2025rdt}
Songming Liu, Lingxuan Wu, Bangguo Li, Hengkai Tan, Huayu Chen, Zhengyi Wang, Ke~Xu, Hang Su, and Jun Zhu.
\newblock Rdt-1b: a diffusion foundation model for bimanual manipulation.
\newblock In \emph{International Conference on Learning Representations}, volume 2025, pages 29982--30009, 2025{\natexlab{b}}.

\bibitem[Liu et~al.(2026)Liu, Li, Ma, Wu, Tan, Ouyang, Su, and Zhu]{liu2026rdt2}
Songming Liu, Bangguo Li, Kai Ma, Lingxuan Wu, Hengkai Tan, Xiao Ouyang, Hang Su, and Jun Zhu.
\newblock Rdt2: Exploring the scaling limit of umi data towards zero-shot cross-embodiment generalization.
\newblock \emph{arXiv preprint arXiv:2602.03310}, 2026.

\bibitem[Liu et~al.(2022)Liu, Gong, and Liu]{liu2022flow}
Xingchao Liu, Chengyue Gong, and Qiang Liu.
\newblock Flow straight and fast: Learning to generate and transfer data with rectified flow.
\newblock \emph{arXiv preprint arXiv:2209.03003}, 2022.

\bibitem[Luo et~al.(2025)Luo, Feng, Zhang, Zheng, Wang, Yuan, Liu, Xu, Jin, and Lu]{luo2025being}
Hao Luo, Yicheng Feng, Wanpeng Zhang, Sipeng Zheng, Ye~Wang, Haoqi Yuan, Jiazheng Liu, Chaoyi Xu, Qin Jin, and Zongqing Lu.
\newblock Being-h0: vision-language-action pretraining from large-scale human videos.
\newblock \emph{arXiv preprint arXiv:2507.15597}, 2025.

\bibitem[Ma et~al.(2026)Ma, Zheng, Wang, Jiang, Cui, Liang, and Yang]{ma2026dit4dit}
Teli Ma, Jia Zheng, Zifan Wang, Chunli Jiang, Andy Cui, Junwei Liang, and Shuo Yang.
\newblock Dit4dit: Jointly modeling video dynamics and actions for generalizable robot control.
\newblock \emph{arXiv preprint arXiv:2603.10448}, 2026.

\bibitem[Nasiriany et~al.(2024)Nasiriany, Maddukuri, Zhang, Parikh, Lo, Joshi, Mandlekar, and Zhu]{nasiriany2024robocasa}
Soroush Nasiriany, Abhiram Maddukuri, Lance Zhang, Adeet Parikh, Aaron Lo, Abhishek Joshi, Ajay Mandlekar, and Yuke Zhu.
\newblock Robocasa: Large-scale simulation of everyday tasks for generalist robots.
\newblock \emph{arXiv preprint arXiv:2406.02523}, 2024.

\bibitem[Nasiriany et~al.(2026)Nasiriany, Nasiriany, Maddukuri, and Zhu]{nasiriany2026robocasa365}
Soroush Nasiriany, Sepehr Nasiriany, Abhiram Maddukuri, and Yuke Zhu.
\newblock Robocasa365: A large-scale simulation framework for training and benchmarking generalist robots.
\newblock \emph{arXiv preprint arXiv:2603.04356}, 2026.

\bibitem[NVIDIA et~al.(2025)NVIDIA, Bjorck, Fernando~Castañeda, Da, Ding, Fan, Fang, Fox, Hu, Huang, Jang, Jiang, Kautz, Kundalia, Lao, Li, Lin, Lin, Liu, Llontop, Magne, Mandlekar, Narayan, Nasiriany, Reed, Tan, Wang, Wang, Wang, Wang, Xiang, Xie, Xu, Xu, Ye, Yu, Zhang, Zhang, Zhao, Zheng, and Zhu]{gr00tn1_2025}
NVIDIA, Johan Bjorck, Nikita~Cherniadev Fernando~Castañeda, Xingye Da, Runyu Ding, Linxi~"Jim" Fan, Yu~Fang, Dieter Fox, Fengyuan Hu, Spencer Huang, Joel Jang, Zhenyu Jiang, Jan Kautz, Kaushil Kundalia, Lawrence Lao, Zhiqi Li, Zongyu Lin, Kevin Lin, Guilin Liu, Edith Llontop, Loic Magne, Ajay Mandlekar, Avnish Narayan, Soroush Nasiriany, Scott Reed, You~Liang Tan, Guanzhi Wang, Zu~Wang, Jing Wang, Qi~Wang, Jiannan Xiang, Yuqi Xie, Yinzhen Xu, Zhenjia Xu, Seonghyeon Ye, Zhiding Yu, Ao~Zhang, Hao Zhang, Yizhou Zhao, Ruijie Zheng, and Yuke Zhu.
\newblock {GR00T} {N1}: An open foundation model for generalist humanoid robots.
\newblock In \emph{ArXiv Preprint}, March 2025.

\bibitem[O’Neill et~al.(2024)O’Neill, Rehman, Maddukuri, Gupta, Padalkar, Lee, Pooley, Gupta, Mandlekar, Jain, et~al.]{o2024open}
Abby O’Neill, Abdul Rehman, Abhiram Maddukuri, Abhishek Gupta, Abhishek Padalkar, Abraham Lee, Acorn Pooley, Agrim Gupta, Ajay Mandlekar, Ajinkya Jain, et~al.
\newblock Open x-embodiment: Robotic learning datasets and rt-x models: Open x-embodiment collaboration 0.
\newblock In \emph{2024 IEEE International Conference on Robotics and Automation (ICRA)}, pages 6892--6903. IEEE, 2024.

\bibitem[Pai et~al.(2025)Pai, Achenbach, Montesinos, Forrai, Mees, and Nava]{pai2025mimic}
Jonas Pai, Liam Achenbach, Victoriano Montesinos, Benedek Forrai, Oier Mees, and Elvis Nava.
\newblock mimic-video: Video-action models for generalizable robot control beyond vlas.
\newblock \emph{arXiv preprint arXiv:2512.15692}, 2025.

\bibitem[Peebles and Xie(2023)]{peebles2023scalable}
William Peebles and Saining Xie.
\newblock Scalable diffusion models with transformers.
\newblock In \emph{Proceedings of the IEEE/CVF international conference on computer vision}, pages 4195--4205, 2023.

\bibitem[Pertsch et~al.(2025)Pertsch, Stachowicz, Ichter, Driess, Nair, Vuong, Mees, Finn, and Levine]{pertsch2025fast}
Karl Pertsch, Kyle Stachowicz, Brian Ichter, Danny Driess, Suraj Nair, Quan Vuong, Oier Mees, Chelsea Finn, and Sergey Levine.
\newblock Fast: Efficient action tokenization for vision-language-action models.
\newblock \emph{arXiv preprint arXiv:2501.09747}, 2025.

\bibitem[Qi et~al.(2025)Qi, Wang, Lin, Yi, Ma, Sreenath, and Malik]{qi2025coordinated}
Haozhi Qi, Yen-Jen Wang, Toru Lin, Brent Yi, Yi~Ma, Koushil Sreenath, and Jitendra Malik.
\newblock Coordinated humanoid manipulation with choice policies.
\newblock \emph{arXiv preprint arXiv:2512.25072}, 2025.

\bibitem[Qu et~al.(2025)Qu, Song, Chen, Yao, Ye, Ding, Wang, Gu, Zhao, Wang, et~al.]{qu2025spatialvla}
Delin Qu, Haoming Song, Qizhi Chen, Yuanqi Yao, Xinyi Ye, Yan Ding, Zhigang Wang, JiaYuan Gu, Bin Zhao, Dong Wang, et~al.
\newblock Spatialvla: Exploring spatial representations for visual-language-action model.
\newblock \emph{arXiv preprint arXiv:2501.15830}, 2025.

\bibitem[Sun et~al.(2026)Sun, Zhang, Yang, Zhao, Li, Guo, Song, Ding, Suo, Su, et~al.]{sun2026revisiting}
Nan Sun, Yuan Zhang, Yongkun Yang, Wentao Zhao, Peiyan Li, Jun Guo, Wenxuan Song, Pengxiang Ding, Runze Suo, Yifei Su, et~al.
\newblock Revisiting embodied chain-of-thought for generalizable robot manipulation.
\newblock \emph{arXiv preprint arXiv:2606.03784}, 2026.

\bibitem[Team et~al.(2023)Team, Anil, Borgeaud, Alayrac, Yu, Soricut, Schalkwyk, Dai, Hauth, Millican, et~al.]{team2023gemini}
Gemini Team, Rohan Anil, Sebastian Borgeaud, Jean-Baptiste Alayrac, Jiahui Yu, Radu Soricut, Johan Schalkwyk, Andrew~M Dai, Anja Hauth, Katie Millican, et~al.
\newblock Gemini: a family of highly capable multimodal models.
\newblock \emph{arXiv preprint arXiv:2312.11805}, 2023.

\bibitem[Team et~al.(2024{\natexlab{a}})Team, Georgiev, Lei, Burnell, Bai, Gulati, Tanzer, Vincent, Pan, Wang, et~al.]{team2024gemini}
Gemini Team, Petko Georgiev, Ving~Ian Lei, Ryan Burnell, Libin Bai, Anmol Gulati, Garrett Tanzer, Damien Vincent, Zhufeng Pan, Shibo Wang, et~al.
\newblock Gemini 1.5: Unlocking multimodal understanding across millions of tokens of context.
\newblock \emph{arXiv preprint arXiv:2403.05530}, 2024{\natexlab{a}}.

\bibitem[Team et~al.(2025)Team, Abeyruwan, Ainslie, Alayrac, Arenas, Armstrong, Balakrishna, Baruch, Bauza, Blokzijl, et~al.]{team2025gemini}
Gemini~Robotics Team, Saminda Abeyruwan, Joshua Ainslie, Jean-Baptiste Alayrac, Montserrat~Gonzalez Arenas, Travis Armstrong, Ashwin Balakrishna, Robert Baruch, Maria Bauza, Michiel Blokzijl, et~al.
\newblock Gemini robotics: Bringing ai into the physical world.
\newblock \emph{arXiv preprint arXiv:2503.20020}, 2025.

\bibitem[Team(2025)]{generalist2025gen0}
Generalist Team.
\newblock Gen-0: Embodied foundation models that scale with physical interaction.
\newblock \emph{Generalist AI Blog}, 2025.
\newblock https://generalistai.com/blog/gen-0.

\bibitem[Team(2026{\natexlab{a}})]{generalist2026gen1}
Generalist Team.
\newblock Gen-1: Scaling embodied foundation models to mastery.
\newblock \emph{Generalist AI Blog}, 2026{\natexlab{a}}.
\newblock https://generalistai.com/blog/gen-1.

\bibitem[Team(2026{\natexlab{b}})]{genesis2026gene265}
Genesis~AI Team.
\newblock Gene-26.5: Advancing robotic manipulation to human level.
\newblock \emph{Genesis AI Blog}, May 2026{\natexlab{b}}.
\newblock URL \url{https://genesis.ai/blog/gene-26-5-advancing-robotic-manipulation-to-human-level}.

\bibitem[Team et~al.(2026)Team, Xiang, Bao, Liu, Tan, Bi, Li, Liu, Pang, Jing, et~al.]{team2026motubrain}
MotuBrain Team, Chendong Xiang, Fan Bao, Haitian Liu, Hengkai Tan, Hongzhe Bi, James Li, Jiabao Liu, Jingrui Pang, Kiro Jing, et~al.
\newblock Motubrain: An advanced world action model for robot control.
\newblock \emph{arXiv preprint arXiv:2604.27792}, 2026.

\bibitem[Team et~al.(2024{\natexlab{b}})Team, Ghosh, Walke, Pertsch, Black, Mees, Dasari, Hejna, Kreiman, Xu, et~al.]{team2024octo}
Octo~Model Team, Dibya Ghosh, Homer Walke, Karl Pertsch, Kevin Black, Oier Mees, Sudeep Dasari, Joey Hejna, Tobias Kreiman, Charles Xu, et~al.
\newblock Octo: An open-source generalist robot policy.
\newblock \emph{arXiv preprint arXiv:2405.12213}, 2024{\natexlab{b}}.

\bibitem[Team(2026{\natexlab{c}})]{qwen35blog}
Qwen Team.
\newblock Qwen3.5: Accelerating productivity with native multimodal agents, February 2026{\natexlab{c}}.
\newblock URL \url{https://qwen.ai/blog?id=qwen3.5}.

\bibitem[Team(2026{\natexlab{d}})]{qwenrobotmanip}
Qwen Team.
\newblock Qwen-robotmanip technical report: Alignment unlocks scale for robotic manipulation foundation models.
\newblock 2026{\natexlab{d}}.

\bibitem[Touvron et~al.(2023)Touvron, Lavril, Izacard, Martinet, Lachaux, Lacroix, Rozi{\`e}re, Goyal, Hambro, Azhar, et~al.]{touvron2023llama}
Hugo Touvron, Thibaut Lavril, Gautier Izacard, Xavier Martinet, Marie-Anne Lachaux, Timoth{\'e}e Lacroix, Baptiste Rozi{\`e}re, Naman Goyal, Eric Hambro, Faisal Azhar, et~al.
\newblock Llama: Open and efficient foundation language models.
\newblock \emph{arXiv preprint arXiv:2302.13971}, 2023.

\bibitem[Vuong et~al.(2026)Vuong, Van~Vo, Sohail, Ding, Ma, Liang, Duan, Laptev, and Reid]{vuong2026world2act}
An~Dinh Vuong, Tuan Van~Vo, Abdullah Sohail, Haoran Ding, Liang Ma, Xiaodan Liang, Anqing Duan, Ivan Laptev, and Ian Reid.
\newblock World2act: Latent action post-training from world model dynamics.
\newblock \emph{arXiv preprint arXiv:2603.10422}, 2026.

\bibitem[Walke et~al.(2023)Walke, Black, Zhao, Vuong, Zheng, Hansen-Estruch, He, Myers, Kim, Du, et~al.]{walke2023bridgedata}
Homer~Rich Walke, Kevin Black, Tony~Z Zhao, Quan Vuong, Chongyi Zheng, Philippe Hansen-Estruch, Andre~Wang He, Vivek Myers, Moo~Jin Kim, Max Du, et~al.
\newblock Bridgedata v2: A dataset for robot learning at scale.
\newblock In \emph{Conference on Robot Learning}, pages 1723--1736. PMLR, 2023.

\bibitem[Wang et~al.(2024)Wang, Zhu, Huang, Wang, Chen, and Lu]{wang2023world}
Xiaofeng Wang, Zheng Zhu, Guan Huang, Boyuan Wang, Xinze Chen, and Jiwen Lu.
\newblock Worlddreamer: Towards general world models for video generation via predicting masked tokens.
\newblock \emph{arXiv preprint arXiv:2401.09985}, 2024.

\bibitem[Wu et~al.(2026)Wu, Lu, Wang, Yang, Liu, Wang, Zhu, Sun, Wang, Ma, et~al.]{wu2026pragmatic}
Wei Wu, Fan Lu, Yunnan Wang, Shuai Yang, Shi Liu, Fangjing Wang, Qian Zhu, He~Sun, Yong Wang, Shuailei Ma, et~al.
\newblock A pragmatic vla foundation model.
\newblock \emph{arXiv preprint arXiv:2601.18692}, 2026.

\bibitem[Xu et~al.(2025)Xu, Zhang, Hou, Xu, Fan, Veloso, and Song]{xu2025dexumi}
Mengda Xu, Han Zhang, Yifan Hou, Zhenjia Xu, Linxi Fan, Manuela Veloso, and Shuran Song.
\newblock Dexumi: Using human hand as the universal manipulation interface for dexterous manipulation.
\newblock \emph{arXiv preprint arXiv:2505.21864}, 2025.

\bibitem[Yang et~al.(2026)Yang, Mu, Wei, Lu, Li, Xu, Xue, Yuan, Lin, Pang, et~al.]{yang2026memorywam}
Sizhe Yang, Juncheng Mu, Tianming Wei, Chenhao Lu, Xiaofan Li, Linning Xu, Zhengrong Xue, Zhecheng Yuan, Dahua Lin, Jiangmiao Pang, et~al.
\newblock Memorywam: Efficient world action modeling with persistent memory.
\newblock \emph{arXiv preprint arXiv:2606.20562}, 2026.

\bibitem[Ye et~al.(2026{\natexlab{a}})Ye, Wang, Ni, Huang, Zhao, Li, Li, Li, Lv, Liu, Cao, Li, Deng, Mei, Wang, Chen, Zhou, Wang, Chang, Li, Zhou, Ye, Liu, and Zhu]{ye2026gigaworld}
Angen Ye, Boyuan Wang, Chaojun Ni, Guan Huang, Guosheng Zhao, Hao Li, Hengtao Li, Jie Li, Jindi Lv, Jingyu Liu, Min Cao, Peng Li, Qiuping Deng, Wenjun Mei, Xiaofeng Wang, Xinze Chen, Xinyu Zhou, Yang Wang, Yifan Chang, Yifan Li, Yukun Zhou, Yun Ye, Zhichao Liu, and Zheng Zhu.
\newblock Gigaworld-policy: An efficient action-centered world-action model.
\newblock \emph{arXiv preprint arXiv:2603.17240}, 2026{\natexlab{a}}.

\bibitem[Ye et~al.(2026{\natexlab{b}})Ye, Gao, Yang, Zheng, Wang, Chen, Chen, Chen, Liu, and Jia]{ye2026starvla}
Jinhui Ye, Ning Gao, Senqiao Yang, Jinliang Zheng, Zixuan Wang, Yuxin Chen, Pengguang Chen, Yilun Chen, Shu Liu, and Jiaya Jia.
\newblock Starvla-$\alpha$: Reducing complexity in vision-language-action systems.
\newblock In \emph{European Conference on Computer Vision (ECCV)}, 2026{\natexlab{b}}.

\bibitem[Ye et~al.(2026{\natexlab{c}})Ye, Ge, Zheng, Gao, Yu, Kurian, Indupuru, Tan, Zhu, Xiang, et~al.]{ye2026world}
Seonghyeon Ye, Yunhao Ge, Kaiyuan Zheng, Shenyuan Gao, Sihyun Yu, George Kurian, Suneel Indupuru, You~Liang Tan, Chuning Zhu, Jiannan Xiang, et~al.
\newblock World action models are zero-shot policies.
\newblock \emph{arXiv preprint arXiv:2602.15922}, 2026{\natexlab{c}}.

\bibitem[Yu et~al.(2026)Yu, Zhang, Liu, Liu, Kang, Li, Shi, Ma, Yang, Pan, et~al.]{yu2026wall}
Ryan Yu, Pushi Zhang, Starrick Liu, Brae Liu, Miracle Kang, Shalfun Li, Lights Shi, Ellie Ma, Ping Yang, Chris Pan, et~al.
\newblock Wall-oss-0.5 technical report.
\newblock \emph{arXiv preprint arXiv:2605.30877}, 2026.

\bibitem[Yuan et~al.(2026)Yuan, Dong, Liu, and Zhao]{yuan2026fast}
Tianyuan Yuan, Zibin Dong, Yicheng Liu, and Hang Zhao.
\newblock Fast-wam: Do world action models need test-time future imagination?
\newblock \emph{arXiv preprint arXiv:2603.16666}, 2026.

\bibitem[Zawalski et~al.(2024)Zawalski, Chen, Pertsch, Mees, Finn, and Levine]{zawalski2024robotic}
Micha{\l} Zawalski, William Chen, Karl Pertsch, Oier Mees, Chelsea Finn, and Sergey Levine.
\newblock Robotic control via embodied chain-of-thought reasoning.
\newblock \emph{arXiv preprint arXiv:2407.08693}, 2024.

\bibitem[Zhang et~al.(2026{\natexlab{a}})Zhang, Xiang, Lin, Huang, Wang, Zhong, Dong, Wu, Rao, Zhang, et~al.]{zhang2026hy}
He~Zhang, Lingzhu Xiang, Haitao Lin, Zeyu Huang, Minghui Wang, Dingyan Zhong, Yubo Dong, Yihao Wu, Yongming Rao, Dongsheng Zhang, et~al.
\newblock Hy-embodied-0.5-vla: From vision-language-action models to a real-world robot learning stack.
\newblock \emph{arXiv preprint arXiv:2606.14409}, 2026{\natexlab{a}}.

\bibitem[Zhang et~al.(2026{\natexlab{b}})Zhang, Li, Zhang, Yang, Luo, Li, Wang, Wang, Shao, Xu, et~al.]{zhang2026native}
Qihang Zhang, Lin Li, Luyao Zhang, Shuai Yang, Yiming Luo, Shuaiting Li, Ruilin Wang, Junke Wang, Jiahao Shao, Gangwei Xu, et~al.
\newblock Native video-action pretraining for generalizable robot control.
\newblock \emph{arXiv preprint arXiv:2607.08639}, 2026{\natexlab{b}}.

\bibitem[Zhang et~al.(2025)Zhang, Xu, Liu, Yu, Li, Gao, Fei, Yin, Wu, Jiang, et~al.]{zhang2025vlabench}
Shiduo Zhang, Zhe Xu, Peiju Liu, Xiaopeng Yu, Yuan Li, Qinghui Gao, Zhaoye Fei, Zhangyue Yin, Zuxuan Wu, Yu-Gang Jiang, et~al.
\newblock Vlabench: A large-scale benchmark for language-conditioned robotics manipulation with long-horizon reasoning tasks.
\newblock In \emph{Proceedings of the IEEE/CVF International Conference on Computer Vision}, pages 11142--11152, 2025.

\bibitem[Zhao et~al.(2026)Zhao, Zhao, Huang, Li, Zhao, and Li]{zhao2026rynnworld}
Haoyu Zhao, Xingyue Zhao, Siteng Huang, Xin Li, Deli Zhao, and Zhongyu Li.
\newblock Rynnworld-4d: 4d embodied world models for robotic manipulation.
\newblock \emph{arXiv preprint arXiv:2607.06559}, 2026.

\bibitem[Zhao et~al.(2025)Zhao, Lu, Kim, Fu, Zhang, Wu, Li, Ma, Han, Finn, et~al.]{zhao2025cot}
Qingqing Zhao, Yao Lu, Moo~Jin Kim, Zipeng Fu, Zhuoyang Zhang, Yecheng Wu, Zhaoshuo Li, Qianli Ma, Song Han, Chelsea Finn, et~al.
\newblock Cot-vla: Visual chain-of-thought reasoning for vision-language-action models.
\newblock In \emph{Proceedings of the Computer Vision and Pattern Recognition Conference}, pages 1702--1713, 2025.

\bibitem[Zhaxizhuoma et~al.(2025)Zhaxizhuoma, Liu, Guan, Jia, Wu, Liu, Wang, Liang, Chen, Zhang, et~al.]{zhaxizhuoma2025fastumi}
Zhaxizhuom Zhaxizhuoma, Kehui Liu, Chuyue Guan, Zhongjie Jia, Ziniu Wu, Xin Liu, Tianyu Wang, Shuai Liang, Pengan Chen, Pingrui Zhang, et~al.
\newblock Fastumi: A scalable and hardware-independent universal manipulation interface with dataset.
\newblock In \emph{Conference on Robot Learning}, pages 3069--3093. PMLR, 2025.

\bibitem[Zhen et~al.(2025)Zhen, Sun, Zhang, Li, Zhou, Du, and Gan]{zhen2025tesseract}
Haoyu Zhen, Qiao Sun, Hongxin Zhang, Junyan Li, Siyuan Zhou, Yilun Du, and Chuang Gan.
\newblock Tesseract: learning 4d embodied world models.
\newblock \emph{arXiv preprint arXiv:2504.20995}, 2025.

\bibitem[Zheng et~al.(2025{\natexlab{a}})Zheng, Li, Wang, Liu, Kang, Feng, Zheng, Zou, Chen, Zeng, et~al.]{zheng2025x}
Jinliang Zheng, Jianxiong Li, Zhihao Wang, Dongxiu Liu, Xirui Kang, Yuchun Feng, Yinan Zheng, Jiayin Zou, Yilun Chen, Jia Zeng, et~al.
\newblock X-vla: Soft-prompted transformer as scalable cross-embodiment vision-language-action model.
\newblock \emph{arXiv preprint arXiv:2510.10274}, 2025{\natexlab{a}}.

\bibitem[Zheng et~al.(2025{\natexlab{b}})Zheng, Liang, Huang, Gao, Daum{\'e}~III, Kolobov, Huang, and Yang]{zheng2025tracevla}
Ruijie Zheng, Yongyuan Liang, Shuaiyi Huang, Jianfeng Gao, Hal Daum{\'e}~III, Andrey Kolobov, Furong Huang, and Jianwei Yang.
\newblock Tracevla: Visual trace prompting enhances spatial-temporal awareness for generalist robotic policies.
\newblock In \emph{International Conference on Learning Representations}, volume 2025, pages 54277--54296, 2025{\natexlab{b}}.

\bibitem[Zhong et~al.(2026)Zhong, Liu, Wei, Xiong, Yao, Liu, and Ren]{zhong2026acot}
Linqing Zhong, Yi~Liu, Yifei Wei, Ziyu Xiong, Maoqing Yao, Si~Liu, and Guanghui Ren.
\newblock Acot-vla: Action chain-of-thought for vision-language-action models.
\newblock \emph{arXiv preprint arXiv:2601.11404}, 2026.

\bibitem[Zhou et~al.(2024)Zhou, Du, Chen, Li, Yeung, and Gan]{zhou2024robodreamer}
Siyuan Zhou, Yilun Du, Jiaben Chen, Yandong Li, Dit-Yan Yeung, and Chuang Gan.
\newblock Robodreamer: Learning compositional world models for robot imagination.
\newblock \emph{arXiv preprint arXiv:2404.12377}, 2024.

\bibitem[Zhu et~al.(2025)Zhu, Yu, Feng, Burchfiel, Shah, and Gupta]{zhu2025unified}
Chuning Zhu, Raymond Yu, Siyuan Feng, Benjamin Burchfiel, Paarth Shah, and Abhishek Gupta.
\newblock Unified world models: Coupling video and action diffusion for pretraining on large robotic datasets.
\newblock \emph{arXiv preprint arXiv:2504.02792}, 2025.

\bibitem[Zitkovich et~al.(2023)Zitkovich, Yu, Xu, Xu, Xiao, Xia, Wu, Wohlhart, Welker, Wahid, et~al.]{zitkovich2023rt2}
Brianna Zitkovich, Tianhe Yu, Sichun Xu, Peng Xu, Ted Xiao, Fei Xia, Jialin Wu, Paul Wohlhart, Stefan Welker, Ayzaan Wahid, et~al.
\newblock Rt-2: Vision-language-action models transfer web knowledge to robotic control.
\newblock In \emph{Conference on Robot Learning}, pages 2165--2183. PMLR, 2023.

\end{thebibliography}

\clearpage
\appendix

\begin{figure}[t]
    \centering
    \includegraphics[width=0.85\linewidth]{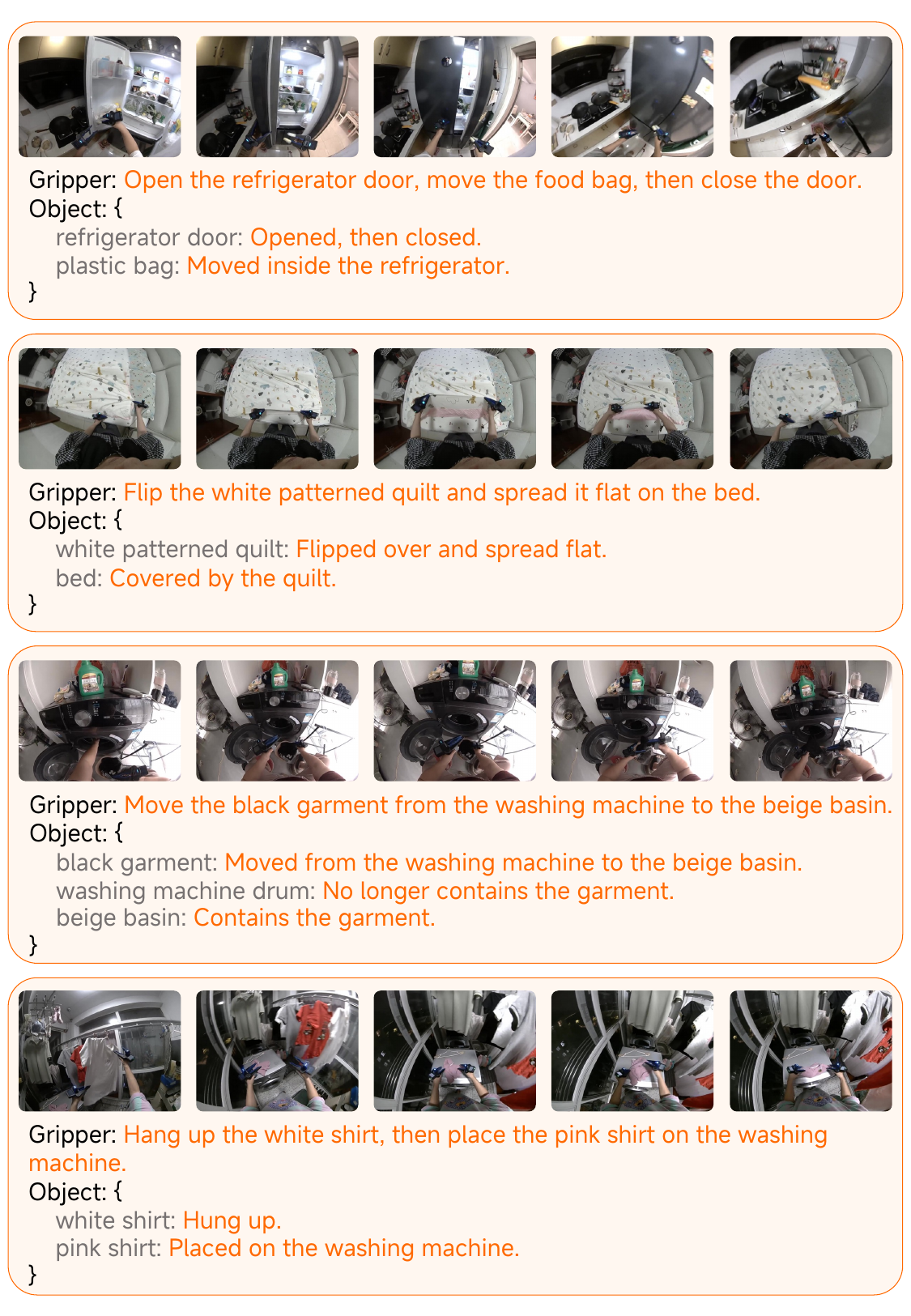}
    \caption{\textbf{Examples of UMI data in the Pre-training Dataset.}}
    \label{fig:appendix_umi_pretrain}
\end{figure}

\begin{figure}[t]
    \centering
    \includegraphics[width=0.85\linewidth]{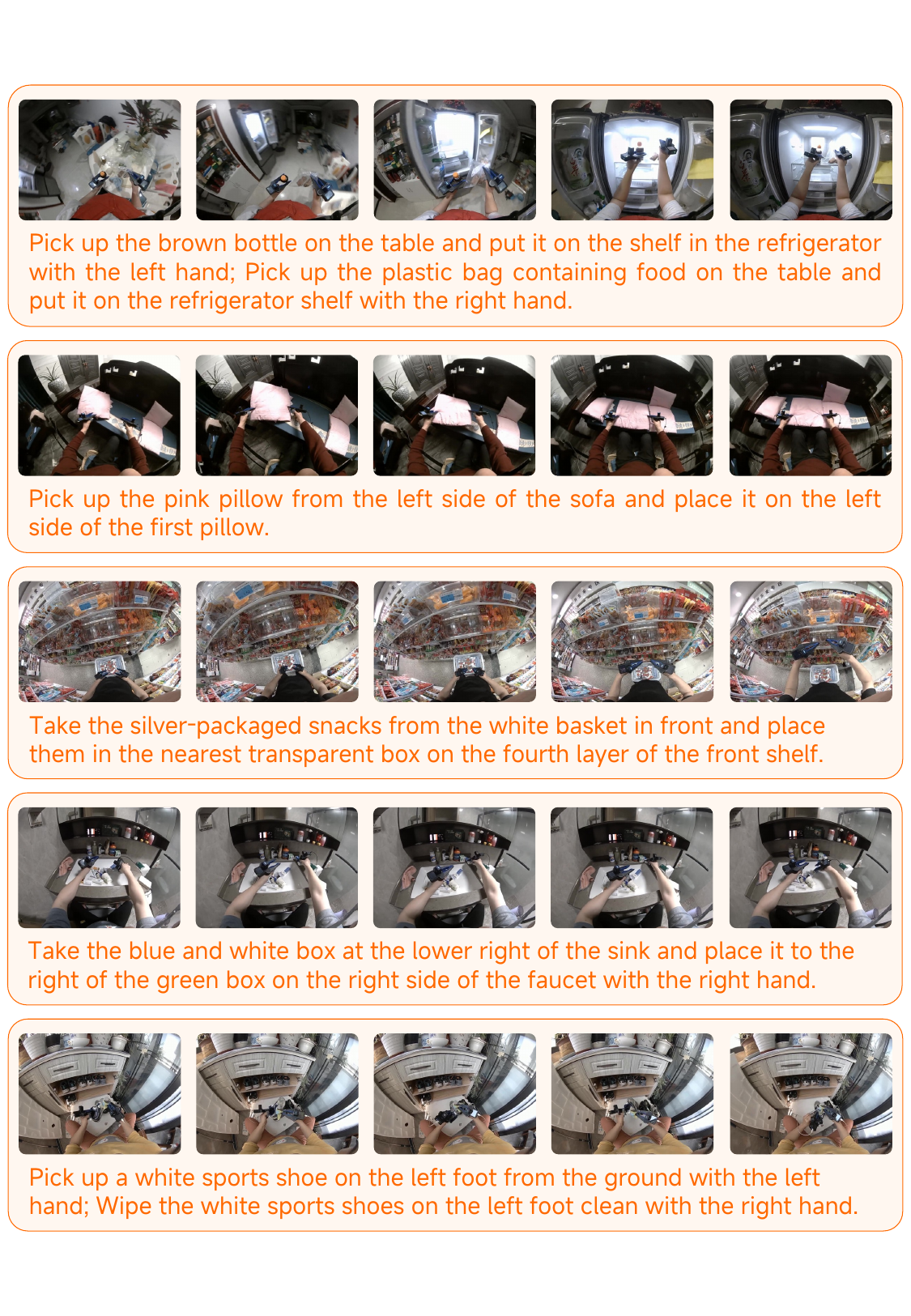}
    \caption{\textbf{Examples of UMI data in the Post-training Dataset.}}
    \label{fig:appendix_umi_posttrain}
\end{figure}

\begin{table*}[t]
    \centering
    \caption{\textbf{Progress Definition for Evaluation on Efficient Adaptation to New Tasks.}
    Each rollout is assigned a progress score from 0 to 100\% according to completed task milestones.}
    \label{tab:finetune_progress}
    \small
    \setlength{\tabcolsep}{4pt}
    \begin{tabularx}{\textwidth}{lXc}
        \toprule
        Task & Progress milestones & Progress (\%) \\
        \midrule
        Phone Packing &
        Grasp the phone; place the phone into the box; grasp the instruction manual; place the manual into the box; grasp the lid; successfully close the lid. &
        10, 10, 30, 10, 10, 30 \\
        \midrule
        Printer Refilling &
        Grasp the paper; complete the handover; successfully insert one end of the paper into the printer tray; fully insert the paper stack into the printer tray; return both robot arms to the resting pose. &
        20, 20, 20, 30, 10 \\
        \midrule
        Laundry Loading &
        Open the washing machine door; move the laundry basket to the door; transfer the clothes into the washing machine; remove the laundry basket; close the washing machine door. &
        20 each \\
        \midrule
        Box Packing &
        Grasp and place each specified target object into the box according to the language instruction. Five target objects are evaluated in each rollout. &
        20 each \\
        \bottomrule
    \end{tabularx}
\end{table*}

\end{document}